\documentclass[sn-vancouver,Numbered, titlepage]{sn-jnl}

\usepackage{graphicx}%
\usepackage{multirow}%
\usepackage{amsmath,amssymb,amsfonts}%
\usepackage{amsthm}%
\usepackage{mathrsfs}%
\usepackage[title]{appendix}%
\usepackage{textcomp}%
\usepackage{manyfoot}%
\usepackage{booktabs}%
\usepackage{algorithm}%
\usepackage{algorithmicx}%
\usepackage{algpseudocode}%
\usepackage{listings}%
\usepackage{longtable}
\usepackage{colortbl}
\usepackage[table]{xcolor} 
\usepackage{epsfig,colortbl}
\usepackage{hhline}
\usepackage{comment}
 \usepackage{booktabs}
\headheight\baselineskip\headsep2\baselineskip\textwidth200mm
\advance\textwidth -1.5in\textheight287mm\advance\textheight-2in
 \label{seq.sub.gdn}
 \usepackage[utf8]{inputenc}
\usepackage{tikz}
\usepackage{amsthm}
\usepackage[singlelinecheck=false]{caption}
\usepackage[font=scriptsize]{caption}
\usepackage[normalem]{ulem}
\usepackage{hyperref}
\usepackage{natbib}
\setcitestyle{numbers,comma,square,sort&compress}
\usepackage{booktabs}
\usepackage{multirow}
\usepackage{graphicx}
\usepackage{array}
\usepackage{amssymb}
\usepackage{changepage} 
\usepackage{amssymb}
\usepackage{ amsthm, amscd, amssymb, graphicx, color,multicol}
 \usepackage{epstopdf}
\usepackage{amssymb}
\usepackage{anyfontsize}
\usepackage{subfigure}
\usepackage{epstopdf}
\usepackage{adjustbox}
\usepackage{lscape}
\usepackage{hyperref}
\usepackage{array}
\usepackage{longtable}
\usepackage{threeparttable} 
\usepackage{changepage}  
\usepackage{adjustbox}
\usepackage{colortbl}  
\usepackage{etoolbox}  
\usepackage{lineno}
\usepackage{geometry}
\usepackage{booktabs}
\usepackage{siunitx}
\usepackage{url,hyperref,lineno,microtype,subcaption}
\usepackage[onehalfspacing]{setspace}
\usepackage{graphicx}%
\usepackage{multirow}%
\usepackage{amsmath,amssymb,amsfonts,amsthm,mathrsfs}%
\usepackage{ amsthm, amscd, amssymb, graphicx, color,multicol}
\usepackage[title]{appendix}%
\usepackage{xcolor}%
\usepackage{textcomp}%
\usepackage{manyfoot}%
\usepackage{booktabs}%
\usepackage{algorithm}%
\usepackage{algorithmicx}%
\usepackage{algpseudocode}%
\usepackage{listings}%
\usepackage{longtable}
\usepackage{colortbl}
\usepackage[table]{xcolor} 
\usepackage{float} 
\usepackage{epsfig,colortbl}
\usepackage{hhline}
\usepackage{comment}
\headheight\baselineskip\headsep2\baselineskip\textwidth200mm
\advance\textwidth -1.5in\textheight287mm\advance\textheight-2in
 \label{seq.sub.gdn}
 \usepackage[utf8]{inputenc}
\usepackage{tikz}
\usepackage[singlelinecheck=false]{caption}
\usepackage[font=scriptsize]{caption}
\usepackage[normalem]{ulem}
\makeatletter
\DeclareRobustCommand{\citebreak}{\unskip\penalty\@highpenalty\space\ignorespaces}
\makeatother
\usepackage{changepage} 
 \usepackage{epstopdf}
\usepackage{subfigure}
\usepackage{epstopdf}
\usepackage{adjustbox}
\usepackage{lscape}
\usepackage{array}
\usepackage{longtable}

\usepackage{changepage}  
\usepackage{adjustbox}
\usepackage{colortbl}  
\usepackage{etoolbox}  
\usepackage{geometry}
\usepackage{setspace}
\usepackage[table]{xcolor}
\usepackage{tabularx}
\usepackage{multirow}
\usepackage{makecell}
\usepackage{array}
\usepackage{adjustbox}
\usepackage{titlesec}
\usepackage{enumitem}

\usepackage{titlesec}
\titleformat{\section}{\normalfont\Large\bfseries}{}{0pt}{}
\author*[1,2]{\fnm{Maryam} \sur{Farhadizadeh}}\email{maryam.farhadizadeh@uniklinik-freiburg.de}

\author[2,3]{\fnm{Maria} \sur{Weymann}}\email{maria.weymann@uniklinik-freiburg.de}

\author[4]{\fnm{Michael} \sur{Blaß}}\email{m.blass@uke.de}

\author[5]{\fnm{Johann} \sur{Kraus}}\email{johann.kraus@uni-ulm.de}
\author[4]{\fnm{Christopher} \sur{Gundler}}\email{c.gundler@uke.de}
\author[6]{\fnm{Sebastian} \sur{ Walter}}\email{swalter@cs.uni-freiburg.de}
\author[1]{\fnm{Noah} \sur{Hempen}}\email{noah.hempen@uniklinik-freiburg.de}
\author[6]{\fnm{Hannah} \sur{Bast}}\email{bast@informatik.uni-freiburg.de}
\author[2,3]{\fnm{Harald} \sur{Binder}}\email{harald.binder@uniklinik-freiburg.de}
\author[1,2]{\fnm{Nadine} \sur{Binder}}\email{nadine.binder@uniklinik-freiburg.de}

\affil*[1]{\orgdiv{Institute of General Practice/Family Medicine}, \orgname{Faculty of Medicine and
Medical Center - University of Freiburg}, \orgaddress{\street{Elsässer Straße 2M}, \city{Freiburg}, \postcode{79110}, \country{Germany}}}

\affil[2]{\orgdiv{Freiburg Center for Data Analysis, Modeling and AI}, \orgname{University of Freiburg}, \orgaddress{\street{Ernst-Zermelo-Str. 1}, \postcode{79104}, \city{Freiburg}, \country{Germany}}}

\affil[3]{\orgdiv{Institute of Medical Biometry and Statistics}, \orgname{Faculty of Medicine and
Medical Center - University of Freiburg}, \orgaddress{\street{Stefan-Meier-Straße 26}, \city{Freiburg}, \postcode{79104}, \country{Germany}}}

\affil[4]{\orgdiv{Institute for Applied Medical Informatics}, \orgname{University Medical Center Hamburg-Eppendorf}, \orgaddress{\street{Martinistraße 52}, \city{Hamburg}, \postcode{20251}, \country{Germany}}}

\affil[5]{\orgdiv{Institute of Medical Systems Biology}, \orgname{Ulm University}, \orgaddress{\street{Helmholtzstraße 16}, \city{Ulm}, \postcode{89081}, \country{Germany}}}

\affil[6]{\orgdiv{Department of Computer Science}, \orgname{Faculty of Engineering - University of Freiburg}, \orgaddress{\street{Georges-Köhler-Allee 051}, \city{Freiburg}, \postcode{79110}, \country{Germany}}}

\title[Challenges and proposed solutions in
modeling multimodal medical data: A systematic
review]{Challenges and proposed solutions in
modeling multimodal medical data: A systematic
review}

\begin{document}

\abstract{
\textbf{Background:} 
Multimodal data analysis has gained significant attention in biomedical Artificial Intelligence. Integrating diverse data sources, such as omics, wearable sensors, imaging, environmental, and electronic health records, offers a more comprehensive view of patient health and supports personalized care.  However, modeling multimodal data poses several challenges that must be addressed to ensure accurate and reliable analysis. This systematic review aims to identify and analyze the challenges encountered in modeling multimodal data in clinical domains.

\textbf{Methods:}
A systematic search of PubMed was conducted to identify original studies published between 2011 and October 2023 that modeled two or more data modalities for medical applications and explicitly addressed at least one modeling challenge. In addition, a supplementary search was performed in Google Scholar to capture further relevant studies. Studies were included if they addressed the use of multimodal data in clinical settings, focusing on data from diverse modalities. Papers that focused solely on multimodal imaging or included audio and video data were excluded during the title and abstract screening. A full-text review of 84 papers was then performed, leading to additional exclusions for studies that did not directly address modeling challenges specific to multimodal data in the clinical domain. Data extraction was performed to identify reported challenges along with their corresponding solutions or proposed approaches.

\textbf{Results:}
This review included 69 studies (58 from the systematic search and 11 identified through a supplementary Google Scholar search). Five major categories of challenges were identified: missing modalities, imbalance in dimensionality, optimal fusion strategy, small datasets, and interpretability. Small datasets (17/69, 25\%) and missing modalities (15/69, 22\%) were the most frequently addressed challenges. Intermediate fusion was the most commonly used fusion strategy (39/69 studies). Missing modalities were addressed using techniques such as multitask learning, matrix completion, and generative adversarial networks. Small datasets were commonly tackled through data augmentation, transfer learning, and knowledge distillation. Interpretability was addressed using attention mechanisms and gradient-based attribution methods, while neural architecture search was explored to optimize fusion strategies.

\textbf{Conclusions:}
Addressing the main challenges of multimodal modeling in clinical domains, this review summarizes emerging strategies and synthesizes current trends and methodological advances. It provides a comprehensive overview of the field, highlights opportunities for future research and development, and aims to support the design of more robust multimodal models and improve data integration for clinical applications.

}

\maketitle
\section*{Background}
\begin{figure}[t]
\centering         
\includegraphics[scale=0.45]{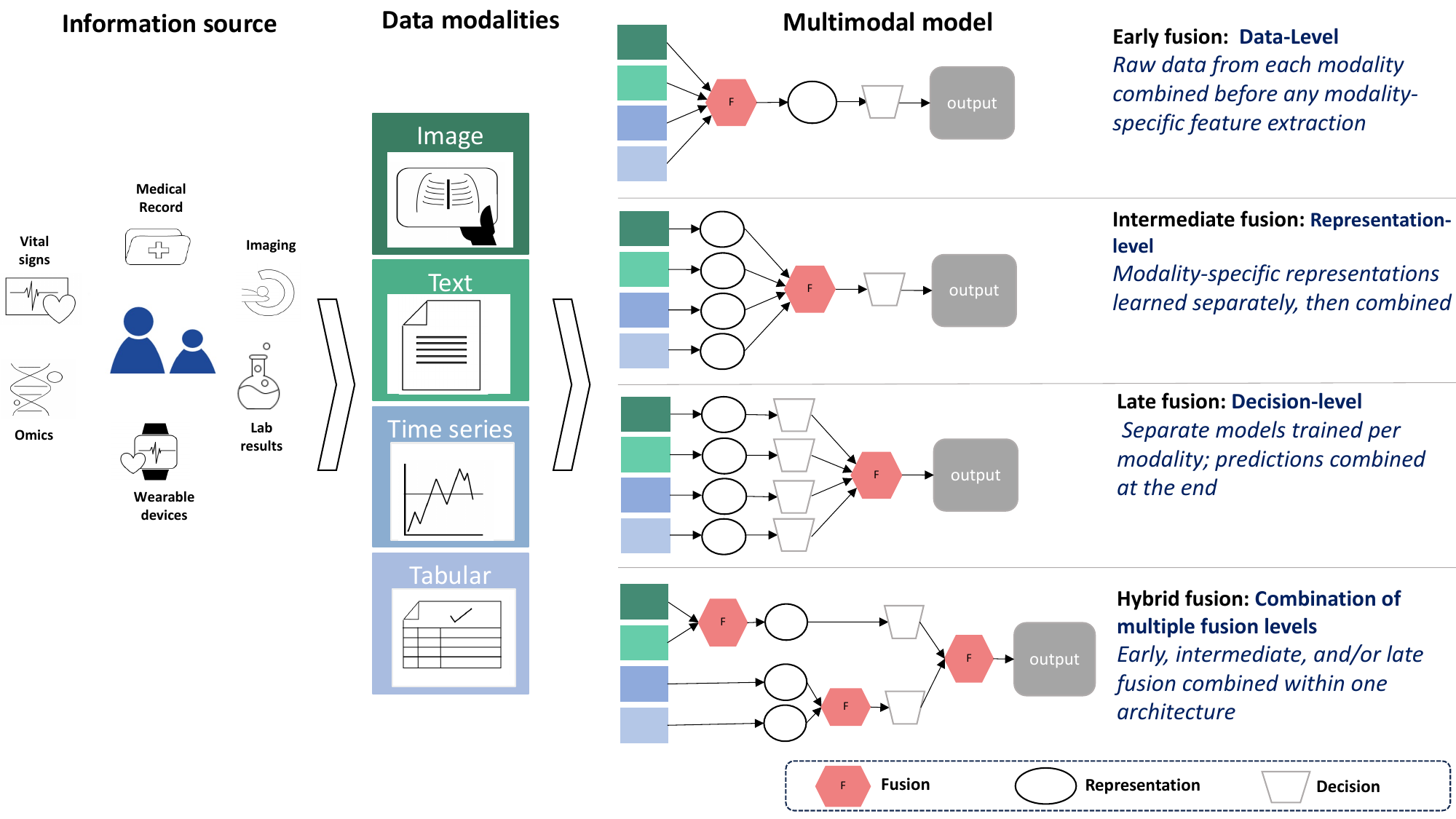}
\captionsetup{skip=13pt}
\caption{Multimodal Modeling: Various information sources can be collected from a patient during a clinical stay or a doctor's visit resulting in different data modalities.  Subsequently, the modalities can be evaluated jointly using a suitable fusion technique, where the fusion (F) can be happening at different stages in the process.}
\label{fig1}
\end{figure}

In healthcare, data comes in various modalities like electronic health records (EHR), imaging, genomics, and wearable devices. Integrating more than one of these modalities in a multimodal data analysis holds significant promise for improving our understanding and management of complex health conditions \cite{mohsen_artificial_2022,baltrusaitis2019multimodal}. One notable domain where multimodal data analysis can make a transformative impact is in the management of chronic diseases such as asthma and chronic obstructive pulmonary disease (COPD) \cite{acosta2022multimodal}. Here, medical imaging data, such as computed tomography (CT), provides insights into lung morphology and pathological changes. Wearable devices monitor physiological parameters and activity levels, offering real-time data on the patient's disease trajectory. Environmental data, encompassing air quality and pollutants, helps identify external factors influencing respiratory conditions. Clinical records and genomic information contribute to personalized treatment strategies, while patient-reported outcomes capture subjective experiences, shaping a patient-centric approach. Combining these complementary modalities provides a more comprehensive understanding of disease and enables the identification of interactions between different aspects of patient health. 
We use this example to illustrate the general potential of multimodal integration in chronic disease management; as discussed in the Limitations, the studies identified in our systematic search were concentrated predominantly in neurology and oncology rather than in COPD or asthma specifically.
Recent bibliometric analyses further demonstrate the rapidly growing interest in multimodal Artificial Intelligence (AI) for healthcare, highlighting multimodal data fusion as an increasingly active research area with applications spanning diagnosis, prognosis, and clinical decision support \cite{Chen2024Bibliometric}. Despite this rapid progress, modeling multimodal data remains a challenging task that extends well beyond the analysis of individual modalities because individual modalities differ in their data types, formats, scales, levels of noise, and interpretability. Moreover, modalities may exhibit varying degrees of completeness, sparsity, and dimensionality, making it challenging to integrate and analyze them together.
Addressing these challenges requires effective multimodal fusion strategies, which combine information from different modalities into a unified representation. Learning such shared representations across heterogeneous modalities has been a central objective of multimodal deep learning since its early development \cite{ngiam2011multimodal}. Fusion strategies can be broadly divided into four main categories: (i) early fusion, (ii) intermediate fusion, (iii) late fusion, and (iv) hybrid fusion.  
(i) Early fusion involves integrating information from different sources at the data level without learning any marginal representations or feature extraction, treating them as unimodal input. 
Early fusion offers simplicity by integrating modalities at the data level, however, it often assumes that all input modalities are directly compatible and equally informative. It can capture basic cross-modal relationships present in raw data, but it typically struggles to model more complex, high-level interactions between modalities. In particular, early fusion may be inadequate when modalities differ in dimensionality, sampling rates, or semantic content, such as combining image data with structured omics data, where aligning and weighting heterogeneous features is non-trivial.
(ii) Intermediate fusion, in contrast, involves learning modality-specific representations to identify correlations within each modality before generating joint representations or making predictions. This approach offers flexibility in selecting different architectures for feature extraction from various modalities \cite{stahlschmidt_multimodal_2022}. By employing distinct architectures and layers for each modality, it can effectively handle heterogeneous data. Additionally, intermediate fusion is resilient to imbalance in dimensionality \cite{stahlschmidt_multimodal_2022} and can accommodate missing modalities without significantly affecting performance \cite{ stahlschmidt_multimodal_2022, lee2021variational}. (iii) Late fusion, on the other hand, trains separate models for each modality and integrates their individual predictions at a later stage. This method is particularly well-suited for managing heterogeneous modalities and addressing imbalance in dimensionality and missing data. However, it may be less effective for modalities with strong correlations, as it does not capture interactions between different data sources.
In addition, in scenarios where specific information holds greater significance, using late fusion offers notable benefits. The model prioritizes the most important information it has, then adds more details from other sources to improve its understanding. 
This prioritization allows the model to integrate complementary information from multiple modalities while emphasizing the most informative sources, thereby enhancing decision-making and prediction performance.
(iv) Hybrid fusion combines elements from the aforementioned methods, allowing for flexible processing across multiple stages. This approach is especially effective in scenarios with diverse data modalities, where early fusion can be used for closely related data types, and later fusion stages facilitate the integration of more distinct modalities. It provides the adaptability needed to meet the specific requirements of multimodal models. 
Figure~\ref{fig1} illustrates how diverse data modalities can be integrated within a multimodal modeling framework. It depicts the four primary fusion strategies (early, intermediate, late, and hybrid), while other less frequently used strategies, such as hierarchical fusion, probability-based fusion, and maximum correlation-based fusion, are not shown for simplicity but are briefly discussed in Section \ref{sec:result}.

Selecting the optimal fusion technique is only one of several challenges in multimodal modeling \cite{pawlowski_effective_2023}. This systematic review aims to identify the broader set of challenges encountered in modeling multimodal data for medical applications and to synthesize the solutions proposed in the reviewed studies. By structuring the review around key challenges such as missing modalities, small datasets, imbalance in dimensionality, interpretability, and fusion strategies, we provide a challenge-oriented perspective that highlights not only the methodological progress achieved but also the gaps that remain.

Several review articles have discussed the application of deep learning to analyze multimodal data. For instance, \citet{mohsen_artificial_2022} explored integrating multiple data modalities with AI models for a variety of applications, including combining medical imaging with data from EHRs. Similarly, \citet{azam_review_2022} and \citet{zhang_advances_2020} solely focused on fusing different imaging modalities. \citet{behrad_overview_2022} and \citet{stahlschmidt_multimodal_2022} primarily concentrated on integrating omics data with other modalities. \citet{huang_fusion_2020} address the fusion of multimodal EHR data using traditional machine learning and deep learning methods. \citet{pawlowski_effective_2023} and \citet{shaik2023survey} focus on fusion techniques and feature extraction methods. 

While these reviews provide valuable methodological overviews, they are generally organized around data modalities, clinical applications, or fusion methodologies. Consequently, they offer limited insight into how recurring methodological challenges influence model development and performance across different clinical settings. Challenges such as missing modalities, small datasets, imbalance in dimensionality, interpretability, and the selection of appropriate fusion strategies frequently arise across disease domains, modality combinations, and modeling frameworks.

In contrast, this review adopts a challenge-oriented perspective and systematically examines the solutions proposed to address these recurring methodological challenges. By synthesizing evidence across diverse clinical applications and multimodal data types, we identify common methodological patterns, recurring limitations, and areas requiring further methodological development. This perspective complements existing modality- and application-oriented reviews and provides practical guidance for researchers facing specific challenges in multimodal medical data modeling.

Accordingly, the aim of this review is to systematically summarize the major challenges encountered in modeling multimodal medical data and to provide an overview of the solutions proposed in the literature to address them. To achieve this objective, we first identify and screen relevant studies through a systematic review process. We then derive a challenge-oriented taxonomy from the included studies and organize the results according to the major methodological challenges encountered in multimodal medical data modeling. For each challenge, we summarize its impact on multimodal modeling, review the solutions proposed in the literature, and synthesize the evidence across studies to identify common methodological trends, remaining limitations, and future research directions.

The remainder of this paper is organized as follows. In Section \ref{Sec:method}, we present the methodology used for this systematic review, including the search strategy, inclusion and exclusion criteria, and data extraction process. Section \ref{sec:result} presents the challenge taxonomy identified in the reviewed studies and synthesizes the solutions proposed to address each challenge. Section \ref{sec:discussion} discusses the key findings, highlights gaps in the literature, and reflects on limitations and future research directions. Finally, the paper concludes with Section \ref{sec:conclusion}, summarizing the main insights and contributions of this review.

\section{Methods}\label{Sec:method}
\subsection*{Search strategy}

To systematically identify relevant articles, a structured search strategy was employed (see Supplementary Table S2). The search was conducted using two major databases, PubMed and a supplementary search from Google Scholar. The time frame for the search was limited until October 2023 for PubMed (search date: 2023/10/09) and until end of 2023 for Google Scholar (search date: 2024/01/02). The strategy was designed to identify studies addressing challenges and methodologies in modeling multimodal data, with a focus on clinical applications. First, studies were required to explicitly mention ``model fusion'', ``data fusion,'' or ``multimodal'' in their title, ensuring relevance to the topic of multimodal data modeling. To ensure comprehensive coverage, the Boolean operator OR was used within each keyword group to combine related terms. Next, the search included terms related to key research purposes, such as ``disease prediction'', ``diagnosis'', ``clinical outcome prediction'', ``modeling'', or ``exacerbation'', to capture studies with clinical applications. Additionally, the search incorporated terms representing diverse data modalities, including ``electronic health records'', ``imaging data'', ``wearable devices'', ``longitudinal data'', ``temporal data'', or ``time-series data'', to ensure coverage of a wide range of multimodal datasets. To further refine the results, the term ``challenges'' was included to focus on studies discussing methodological or analytical obstacles related to multimodal data integration. Finally, the Boolean operator AND was used to combine these components. The final search query combined these elements to ensure a comprehensive yet focused selection of studies. 
In addition, a supplementary search was performed to identify any potentially relevant studies not captured in the database search.
The reporting of this review follows the PRISMA 2020 guidelines \cite{page2021prisma}.

\subsection*{Inclusion and exclusion criteria} 

Two authors, MF and MW, screened the titles and abstracts of all identified studies to determine their relevance according to prespecified inclusion and exclusion criteria. We selected studies that specifically addressed challenges in the modeling of multimodal data.
Eligible studies were required to explicitly describe, evaluate, or propose a methodological solution to at least one challenge related to multimodal data modeling. Studies that merely applied multimodal models without discussing or addressing such methodological challenges were excluded.
Study publications were included if they focused on methodologies, approaches, or frameworks for integrating, fusing, or analyzing data from multiple modalities. Studies were excluded if they focused exclusively on multimodal imaging or utilized only audio and video data, 
as this review focused on the integration of heterogeneous multimodal medical data (e.g., imaging, clinical records, genomics, laboratory measurements, and wearable data), rather than challenges specific to combining multiple imaging modalities.
Review articles were excluded to maintain an emphasis on original research.
We also excluded publications addressing clinical management, therapeutic interventions, or non-technical challenges (e.g., treatment modalities, surgical techniques, or clinical guidelines) without discussing modeling challenges in multimodal data fusion. Similarly, studies focusing on a single data modality (e.g., imaging, clinical data, or sensor data) without addressing challenges in combining multiple data sources were excluded. Publications centered on clinical applications, such as diagnosis, treatment protocols, or medical decision-making, were included only if they also explored computational modeling or challenges related to multimodal data integration.
Additionally, case studies, clinical trial results, and studies describing specific treatments (e.g., surgical procedures, medical devices, or drug interventions) were excluded unless they explicitly addressed how multimodal data were modeled or fused to improve decision-making, diagnosis, or treatment. 
Finally, although ``multimodal'' was used as a keyword in the search strategy and required to be present in the title, some publications were excluded because their focus was unrelated to data fusion or computational modeling (e.g., multimodal therapy or multimodal plasma), thus falling outside the scope of this review.

The screening process was conducted by five co-authors. The selected full-text studies were strategically distributed among three parties to ensure that each study was independently reviewed twice. Specifically, the first party (MF) screened the first two-thirds, the second party (MB, CG) handled the last two-thirds, and the third party (JK) examined the first and last thirds. In cases of discrepancies, an independent third screener (MW) was brought in to make a final decision.

\subsection*{Data extraction}
For the data extraction process, we categorized the included publications according to the specific challenges they addressed. We identified five main technical challenges in modeling multimodal data: (i) missing modalities and incomplete datasets, (ii) Small datasets (data sparsity), (iii) interpretability, (iv) imbalance in dimensionality, and (v) identifying the optimal fusion strategy. We extracted information on the specific challenges each study tackled within the broader scope of the five identified challenges. We also captured the proposed solutions for overcoming these challenges. We further specifically collected information about the fusion strategies employed in each study. The choice of fusion technique directly relates to the challenge of finding the optimal fusion strategy (challenge v). Some strategies may also help mitigate other challenges, such as handling incomplete datasets (challenge i) or balancing dimensionality (challenge iv) by integrating information from multiple sources effectively. 

To better understand which data types are most frequently integrated, we extracted the modality combinations reported in each study. We also recorded the type of analysis task (e.g., classification, clustering, survival prediction, ...), as this directly determines the methodological setting in which multimodal data are analyzed. In addition, we noted whether studies reported quantitative performance metrics or comparative benchmarks against baseline methods. Capturing this information (e.g., accuracy, area under the curve (AUC), concordance index) was important because consistent reporting of evaluation measures is essential for assessing methodological impact and enabling meaningful comparisons across studies. 
For each included study, we additionally recorded whether a publicly available benchmark dataset (e.g., ADNI, TCGA)  was used for evaluation and whether the authors reported an open-source code implementation (e.g., via a GitHub repository). Each recorded item was verified against the original publication.
In addition, because different medical settings encounter distinct modeling challenges, we recorded the clinical domain of each study to link it with the challenges addressed. 


Findings were synthesized using summary tables (challenge categories, solutions, and modality--task combinations; Tables~\ref{tab:table1} and~\ref{tab:table2}) and visualizations generated in R: bar charts for fusion strategy (Figure~\ref{fig:fig3}), yearly challenge trends (Figure~\ref{fig:fig5}), and modality/task distribution (Figure~\ref{fig:fig6}), stacked dot plot for medical domains (Figure~\ref{fig:fig4}), and a cross-tabulation for benchmark dataset and code availability (Figure~\ref{fig:fig7}). Detailed study-level data are provided in Appendix Table A2 and Supplementary Table S1.

\section{Results}\label{sec:result}

\begin{figure}[t]
\centering  
\includegraphics[width=\textwidth]{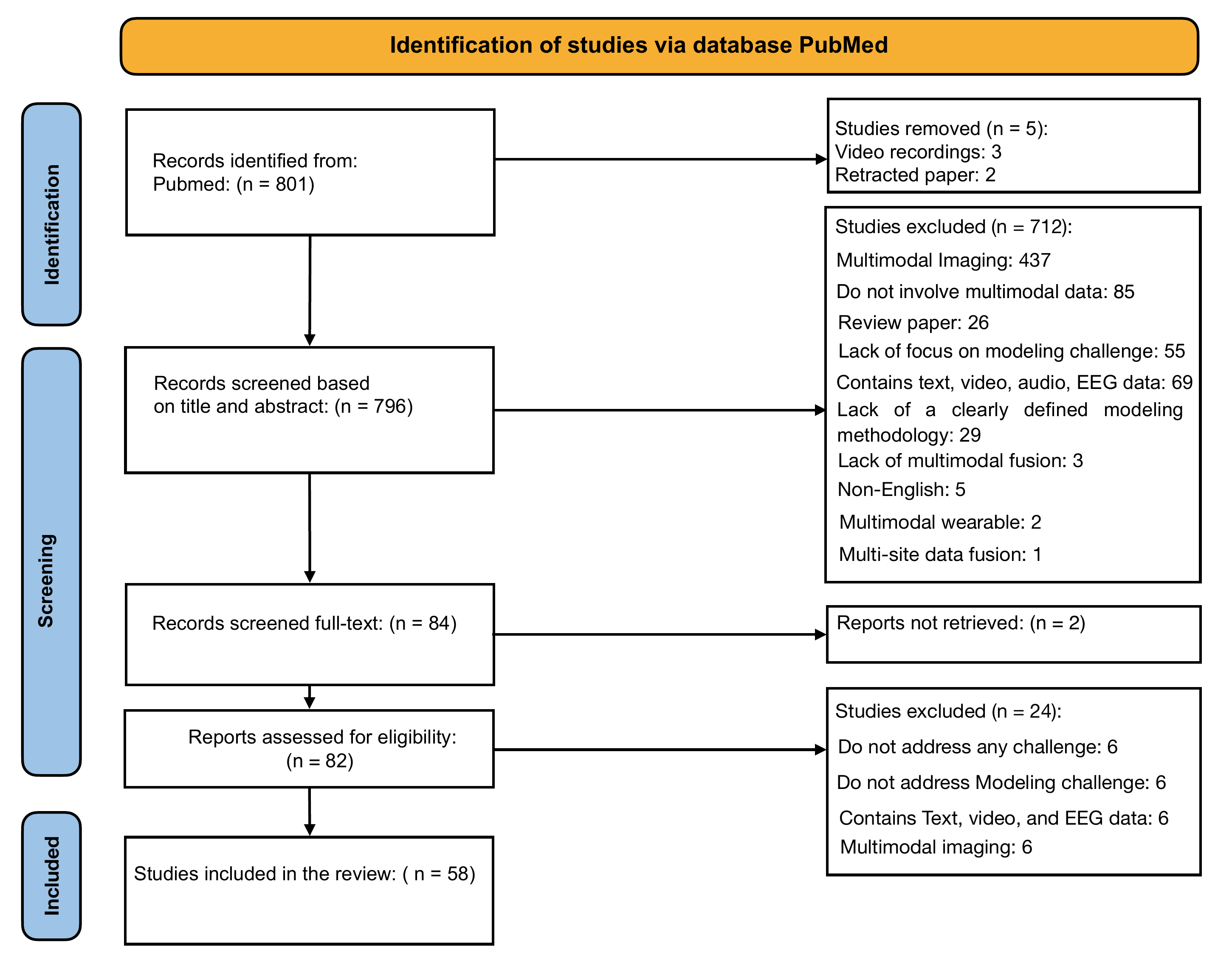}
\caption{PRISMA flow diagram illustrating the study selection process following the systematic search in PubMed. In addition, 11 relevant studies were identified through a supplementary search from Google Scholar, resulting in a total of 69 included studies.}
\label{fig2}
\end{figure}

The initial search yielded a total of 801 publications obtained from PubMed after removing duplicates. Following title and abstract screening, 84 articles met the initial inclusion criteria and were selected for full-text assessment. After full-text review, 58 studies that specifically addressed challenges related to the modeling of multimodal data in medical domains were retained for analysis, while studies that did not directly focus on these modeling challenges or instead addressed clinical challenges were excluded. In addition, 11 relevant studies were identified through a supplementary search from Google Scholar, resulting in a final set of 69 included studies (Figure~\ref{fig2}).

\begin{figure}[t]
\centering         
\includegraphics[scale=0.6]{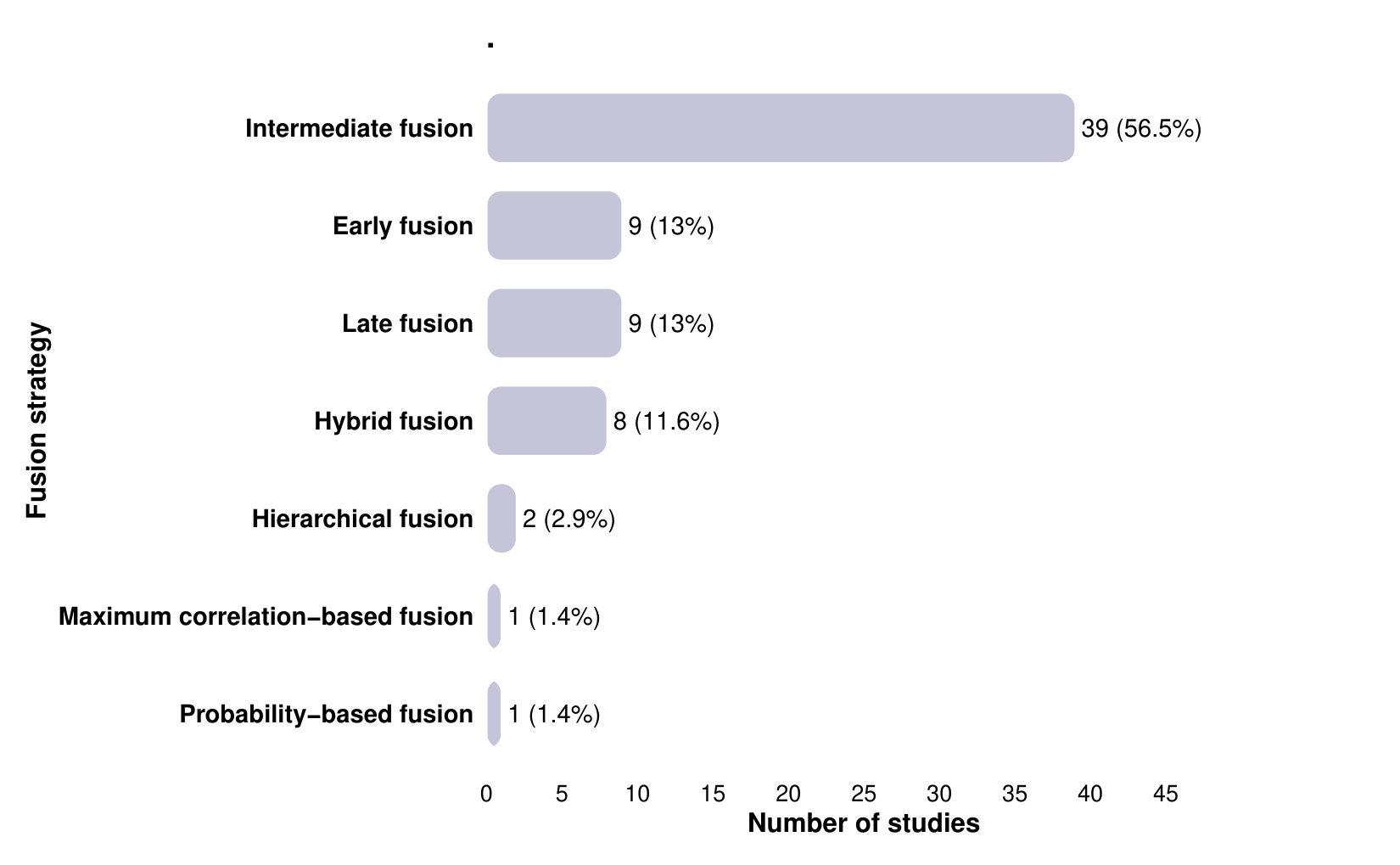}
\caption{Distribution of fusion strategies among the 69 included studies. Bars show the number of studies using each fusion strategy, with counts and percentages labeled directly on each bar. Key takeaway: intermediate fusion was the most frequently used strategy, followed by early and late fusion approaches.}
\label{fig:fig3}
\end{figure}

We summarized our findings in Figure~\ref{fig:fig3}, which provides a comprehensive overview of the fusion strategies within the literature.

Intermediate fusion was the most frequently used strategy, reported in 39 studies, followed by early and late fusion (9 studies) and hybrid fusion (8 studies). Hierarchical fusion was employed in two studies, whereas probability-based fusion and maximum correlation-based fusion were each reported in a single study. 

The two studies employing hierarchical fusion addressed multimodal integration using different architectures. One study 
introduced a transformer-based model that progressively integrates features from shallow to deep layers using hierarchical multimodal transformers. This design facilitates effective cross-modality interaction and addresses issues such as spatial misalignment and data heterogeneity, distinguishing it from traditional early, intermediate, or late fusion techniques. Similarly, one study 
proposed a hierarchical factorized bilinear fusion strategy tailored for cancer survival prediction, which integrates genomic and pathological image data across low- and high-level fusion stages. While both approaches highlight the critical role of cross-modality interactions, the latter further incorporates modality-specific and cross-modality attentional factorized bilinear modules, reducing computational complexity while improving performance, particularly for genomic data analysis. Together, these studies illustrate the flexibility of hierarchical fusion frameworks for integrating heterogeneous medical data.
One study 
employed a probability-based fusion approach, which involves calculating the probability of each feature being relevant and discriminative for classification, as determined by two optimization algorithms. Features with higher probabilities are prioritized and included in the final fused feature set, ensuring the combination of only the most informative and non-redundant features for classification. One other study 
used the maximum correlation-based fusion approach, which emphasizes integrating features with the strongest cross-modal correlations. In this method, features from different modalities are analyzed for their pairwise correlations. Strongly correlated features are selected and combined, while weakly correlated or redundant features are excluded. This strategy enhances the representational power of the fused data by focusing on meaningful relationships while reducing noise and redundancy.

\begin{figure}[t]
\centering   
\hspace*{-1.5cm}
\includegraphics[scale=0.7]{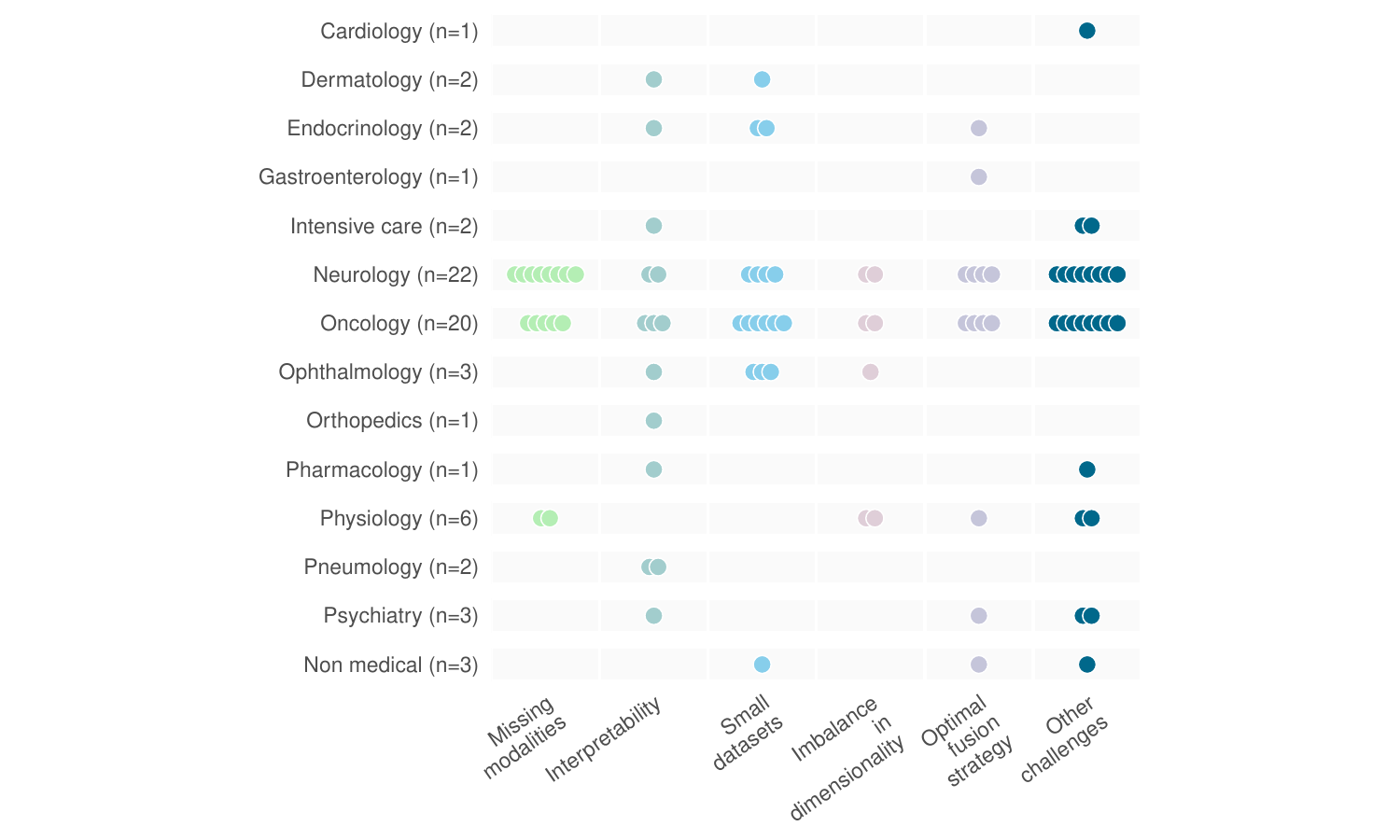}
\caption{Distribution of modeling challenges across medical domains among the 69 included studies. Each dot represents one study; dots are stacked within a domain–challenge cell to indicate the number of studies addressing that challenge within that domain. A single study may address multiple challenges, so the number of studies shown across a domain's row can exceed that domain's total study count (n, given in parentheses). Key takeaway: neurology (n=22) and oncology (n=20) were the most frequently studied domains and also showed the greatest diversity of challenges addressed, particularly missing modalities and small datasets.}
\label{fig:fig4}
\end{figure}


Figure~\ref{fig:fig4} provides an overview of the number of publications across different medical domains, along with the distribution of modeling challenges within these domains. This classification highlights the primary areas of application for multimodal data modeling, with neurology emerging as the most frequently studied domain (22 studies), followed by oncology (20 studies), physiology (6 studies), and ophthalmology and psychiatry (3 studies each). Additional domains such as psychology, endocrinology, cardiology, orthopedics, pharmacology, and gastroenterology further demonstrate the adaptability of multimodal data modeling across various medical contexts. Additionally, three studies explored non-medical multimodal applications. 
Notably, neurology and oncology stand out as the domains most affected by modeling challenges, where missing modalities and small datasets are especially frequent due to the difficulty of integrating heterogeneous data sources (e.g., imaging, genomics, and clinical notes) and the limited availability of samples for rare or complex conditions.

\begin{figure}[t]
\centering         
\includegraphics[scale=0.6]{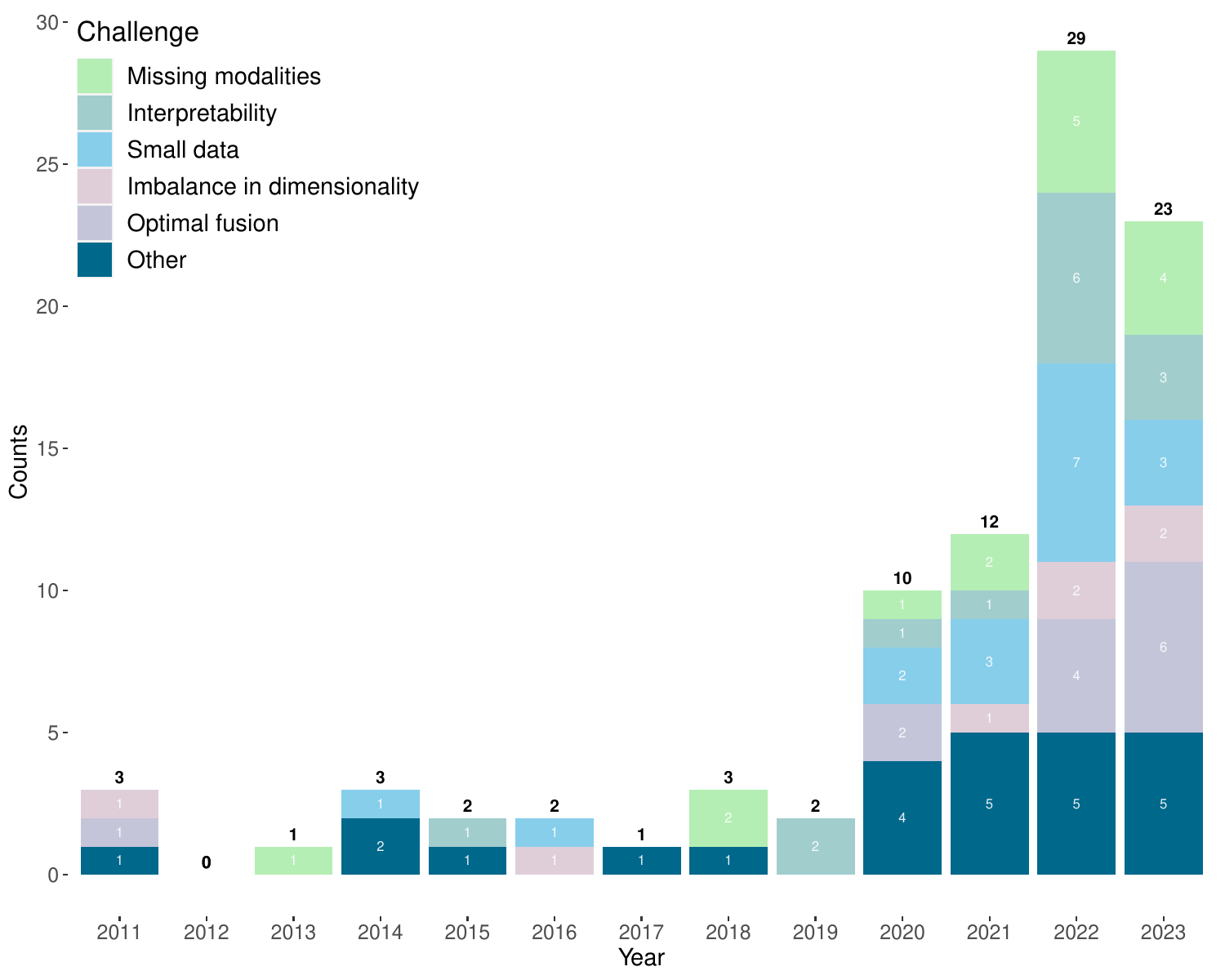}
\caption{ Distribution of key challenges addressed in yearly publications. Stacked bars show the number of studies published each year, with colors indicating the challenges addressed. Because individual studies may address multiple challenges, a single study may contribute to more than one challenge category. Key takeaway: Since 2020, there has been a marked increase in the number of studies addressing all major challenges, particularly small datasets, interpretability, and missing modalities.}
\label{fig:fig5}
\end{figure}

The yearly frequencies of the identified challenges are illustrated in \autoref{fig:fig5}. Specifically, since 2020, there was a strong increase in studies addressing challenges in modeling multimodal data (the number of studies in 2023 is lower as the search strategy was limited until October 2023). This upward trend indicates that research in this area will likely continue to expand, addressing new challenges as multimodal applications become more widespread.

For each study, the particular challenges tackled and corresponding strategies are reported in Supplementary Table S3; these are discussed in more detail, organized by challenge, in the following section. Further explanation of the proposed solutions for each challenge, including the medical domain, disease type, and fusion strategy used in each publication, can be found in Supplementary Table S1 for detailed per-study entries.
 
\begin{figure}[t]
\centering         
\includegraphics[scale=0.5]{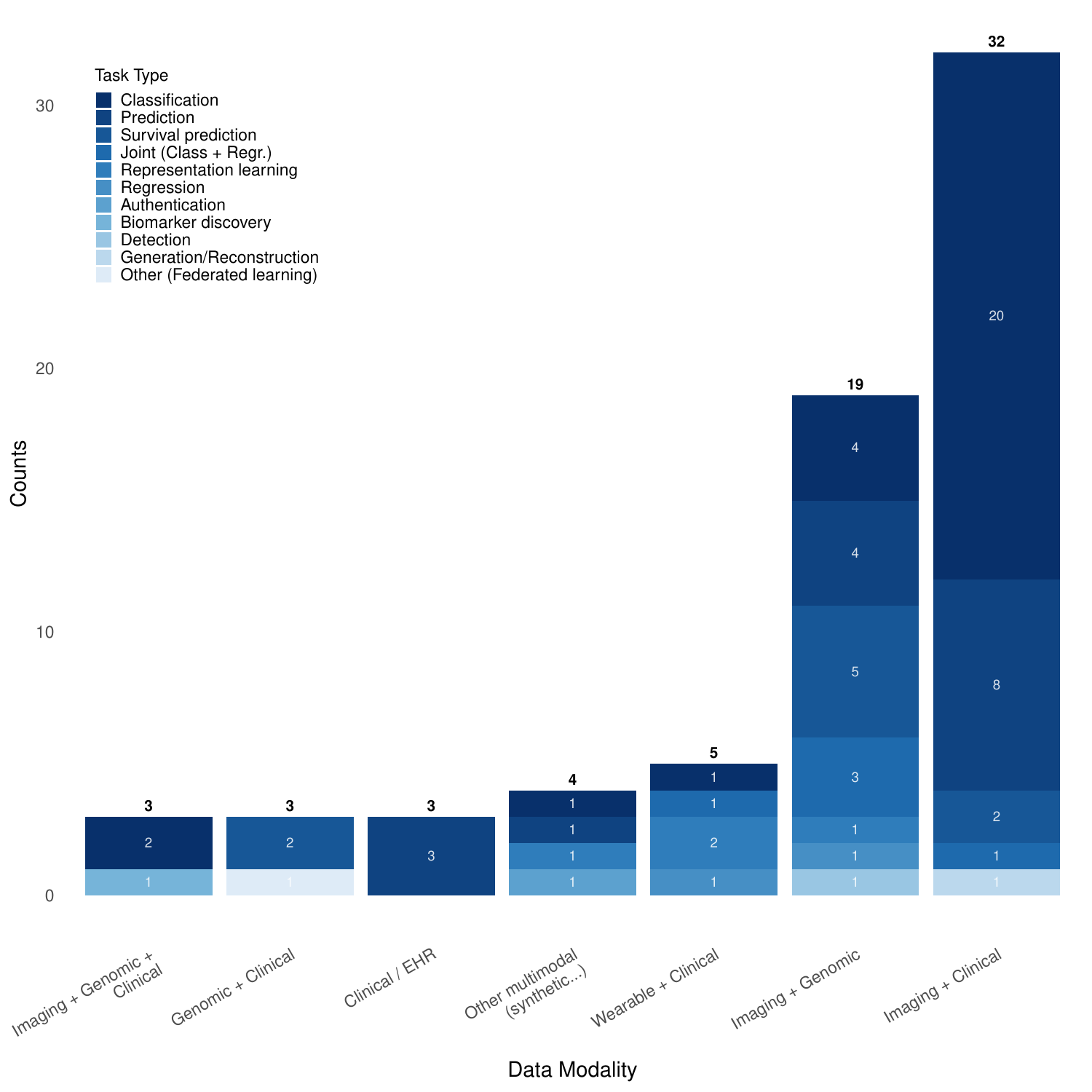}
\caption{Distribution of analysis tasks across data modality combinations. Bars show the number of studies per data modality, subdivided by task type.}
\label{fig:fig6}
\end{figure}

Building on the challenges and solutions reported in Supplementary Table S3, we next analyzed which combinations of data modalities were most frequently investigated and how they relate to the type of task addressed. Table~\ref{tab:table2} and Figure~\ref{fig:fig6} summarize these distributions. The majority of studies combined imaging with clinical data (22 studies) or imaging with genomic data (19 studies), reflecting their central role in current multimodal applications. These combinations were most often applied to classification tasks, with 20 studies using imaging + clinical data and 4 studies using imaging + genomic data, contributing to a total of 27 classification studies across all modality categories. In contrast, other modality pairings were far less common: for example, wearables combined with clinical data appeared in 5 studies, while multi-triple combinations such as imaging + genomic + clinical were reported in only 3 studies.

\definecolor{darkseagreen2}{RGB}{180, 238, 180} 
\definecolor{darkslategray3}{RGB}{162, 205, 205} 
\definecolor{lightskyblue3}{RGB}{135, 206, 235} 
\definecolor{lightsalmon2}{RGB}{248, 152, 128} 
\definecolor{lightpink2}{RGB}{223, 206, 216} 
\definecolor{deepskyblue4}{RGB}{0, 104, 139} 
\definecolor{lightlavender}{RGB}{197, 197, 218} 


\begin{table*}[t]
\centering
\small
\renewcommand{\arraystretch}{1.35}
\setlength{\tabcolsep}{6pt}

\caption{Summary of the major challenges identified in the reviewed studies and representative solution approaches. Key takeaway: Missing modalities and small datasets were the most frequently addressed challenges, while a diverse range of methods has been proposed for each challenge category. Studies addressing challenges outside these five predefined categories are grouped as ``Other Challenges'' and are discussed separately in the text and in Supplementary Table S3, rather than summarized here given their heterogeneity. A detailed study-level mapping of challenges, solutions, and corresponding references is also provided in Supplementary Table S3.}
\label{tab:table1}
\begin{tabularx}{\textwidth}{
c
>{\raggedright\arraybackslash\bfseries}p{2.8cm}
>{\raggedright\arraybackslash}X}
\toprule
& \textbf{Challenge} & \textbf{Representative Methods} \\
\midrule

{\colorbox{darkseagreen2}{\rule{0pt}{4pt}\rule{4pt}{0pt}}} & Missing modalities &
Matrix completion \cite{thung_identification_2013,vivar_multi-modal_2018};
variational autoencoders and latent representation learning \cite{marti-juan_mc-rvae_2023,xu_explainable_2021,xu_multi-modal_2022};
generative adversarial networks \cite{cai_deep_2018,ziegler_multi-modal_2022};
knowledge distillation and knowledge-based approaches \cite{wang_multimodal_2020,xing_gradient_2023};
graph-based methods \cite{hou_hybrid_2023,dong_high-order_2021};
deep learning and boosting approaches \cite{abdelaziz_alzheimers_2021,aghili_addressing_2022,saad_learning-based_2022};
transfer-learning-based handling of missing modalities \cite{sukei_automatic_2023};
attention-based masking mechanisms \cite{vanguri_multimodal_2022}. \\

\addlinespace[0.6em]

{\colorbox{lightskyblue3}{\rule{0pt}{4pt}\rule{4pt}{0pt}}} & Small datasets &
Data augmentation and synthetic data generation \cite{jabeen_breast_2022,partin_data_2023,saad_learning-based_2022};
transfer learning \cite{kustowski_transfer_2021,lim_use_2022,shickel_deep_2021,yoo_deeppdt-net_2022,li_transfer_2023};
knowledge distillation \cite{guan_mri-based_2021};
multi-task learning \cite{liu_prediction_2020};
feature selection, dimensionality reduction, and kernel-learning approaches \cite{lei_discriminative_2016,liu_multiple_2014};
variational autoencoder approaches \cite{xu_explainable_2021};
clustering-based learning \cite{bi_multimodal_2020};
transformer-based architectures \cite{cai_multimodal_2022,cai_fdtrans_2023};
multimodal AI synthesis approaches \cite{boehm_harnessing_2022}. \\

\addlinespace[0.6em]

{\colorbox{darkslategray3}{\rule{0pt}{4pt}\rule{4pt}{0pt}}} & Interpretability &
SHapley Additive exPlanations (SHAP)-based explanations \cite{rabinovici-cohen_multimodal_2022,sheu_multi-modal_2022,wang_interpretability-based_2022,xu_explainable_2021};
Grad-CAM and gradient-based attribution methods \cite{li_transfer_2023,tiulpin_multimodal_2019};
integrated gradients and attention-based explanations \cite{krix_multigml_2023,rahaman_deep_2023};
interpretable neural network architectures \cite{hao_interpretable_2019,xin_interpretation_2021};
feature-importance ranking approaches \cite{li_integrating_2022};
transparent multimodal prediction models \cite{liu_prediction_2020};
guided topic-prior approaches \cite{ahuja_mixehr-guided_2022};
kernel-based interpretable methods \cite{eshaghi_classification_2015}. \\

\addlinespace[0.6em]

{\colorbox{lightpink2}{\rule{0pt}{4pt}\rule{4pt}{0pt}}} & Imbalance in dimensionality &
Feature selection and dimensionality reduction \cite{lei_discriminative_2016,madabhushi_computer-aided_2011,liu_identification_2020};
weighted loss functions \cite{karaman_machine_2022,partin_data_2023};
focal loss functions \cite{mao_cross-modality_2021,li_transfer_2023};
recurrent neural network architectures \cite{xu_multi-modal_2022};
classification-based reweighting approaches \cite{tran_raboc_2016}. \\

\addlinespace[0.6em]

{\colorbox{lightlavender}{\rule{0pt}{4pt}\rule{4pt}{0pt}}} & Optimal fusion strategy &
Attention-based fusion \cite{rahaman_deep_2023,vanguri_multimodal_2022};
hierarchical fusion \cite{li_hfbsurv_2022};
latent representation learning \cite{ning_multi-constraint_2023};
adaptive fusion \cite{xia_graph_2023};
transformer-based fusion \cite{dai_-janet_2023,zhang_tformer_2023};
weighted ensemble fusion \cite{hatami_cnn-lstm_2022};
graph-based embedding methods \cite{tiwari_multi-modal_2011};
canonical-correlation-based fusion \cite{von_luhmann_improved_2020};
knowledge-infused fusion \cite{wang_knowledge-infused_2022};
Gibbs-measure-based fusion \cite{wang_towards_2023};
probability-based fusion \cite{jabeen_breast_2022};
maximum-correlation fusion \cite{khan_stomachnet_2020}. \\

\bottomrule
\end{tabularx}
\end{table*}

\definecolor{groupgray}{gray}{0.92}

\begin{figure}[t]
\centering         
\includegraphics[scale=0.5]{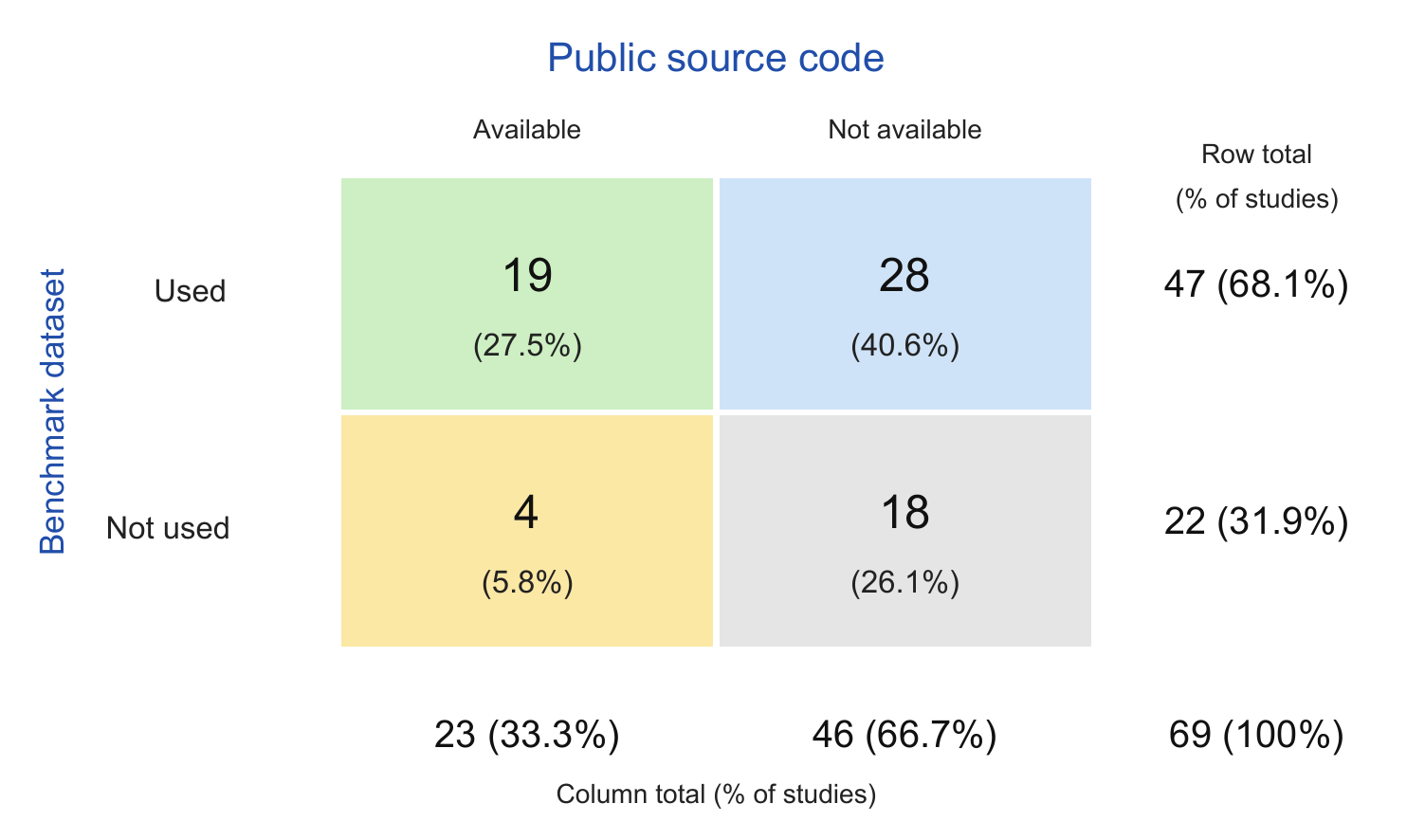}\captionsetup{skip=7pt}
\caption{Cross-tabulation of publicly available benchmark dataset use and public source code availability across the 69 included studies. Row and column totals indicate the number and percentage of studies using a benchmark dataset and reporting public source code, respectively, independent of the other variable. Key takeaway: only 19 studies (27.5\%) reported both a publicly available benchmark dataset and public source code, while 18 studies (26.1\%) lacked both a publicly available benchmark dataset and public source code, limiting opportunities for independent verification and reuse of the underlying implementations.}
\label{fig:fig7}
\end{figure}

\begin{table}[ht]
\centering
\caption{Studies grouped by data modalities, task type, and summarized performance metrics. Key takeaway: Imaging-based multimodal combinations were the most frequently studied and were applied across a broad range of prediction, classification, and survival analysis tasks.}
\label{tab:table2}
\footnotesize
\renewcommand{\arraystretch}{1.2}
\setlength{\tabcolsep}{6pt}
\begin{tabularx}{\textwidth}{|p{0.18\textwidth}|p{0.366\textwidth}|p{0.366\textwidth}|}

\hline
\textbf{Data Modality} & \textbf{Task (metric) + Studies} & \textbf{Performance Metrics} \\
\hline

\rowcolor{groupgray}\multicolumn{3}{|l|}{\textbf{Imaging + Genomic}} \\
& Classification (AUC / Acc) \cite{cai_deep_2018,cai_fdtrans_2023,madabhushi_computer-aided_2011,rahaman_deep_2023} 
& AUC $\approx$ 0.93 \cite{cai_deep_2018,cai_fdtrans_2023}; Acc= 0.90--0.92 \cite{madabhushi_computer-aided_2011}, \cite{rahaman_deep_2023} \\
&  Survival prediction (C-index / AUC) \cite{li_hfbsurv_2022,li_integrating_2022,ning_multi-constraint_2023,shao_fam3l_2023,shi_attention-based_2023} 

& C-index= 0.72–0.78 \cite{ning_multi-constraint_2023,shao_fam3l_2023}; AUC= 0.71–0.77 \cite{shi_attention-based_2023}; C-index= 0.766 \cite{li_hfbsurv_2022,li_integrating_2022} \\
& Prediction \cite{krix_multigml_2023, partin_data_2023, saad_learning-based_2022, vanguri_multimodal_2022}

& AUC= 0.85 \cite{saad_learning-based_2022}; AUC= 0.80 \cite{vanguri_multimodal_2022}; \newline AUROC= 0.898 \cite{krix_multigml_2023}; AUROC= 0.806 \cite{partin_data_2023} \\
& Joint (Class. + Regr.) \cite{abdelaziz_alzheimers_2021, brand_joint_2020}
  
& RMSE= 10 \cite{abdelaziz_alzheimers_2021}; RMSE= 16.2 ± 0.76 \cite{brand_joint_2020} \\
& Representation learning \cite{bi_multimodal_2020} 
& Acc= 0.90 \\
& Regression (RMSE) \cite{zille_enforcing_2017} 
& Not reported \\
& Detection \cite{alam_kernel_2018} 
& $p$-value=0.000001 \\
\hline

\rowcolor{groupgray}\multicolumn{3}{|l|}{\textbf{Imaging + Clinical}} \\

& \raggedright Classification (AUC / Acc) 
\cite{aghili_addressing_2022, cai_multimodal_2022, dai_-janet_2023, dong_high-order_2021,eshaghi_classification_2015, guan_mri-based_2021,jabeen_breast_2022, khan_stomachnet_2020, lei_discriminative_2016, li_transfer_2023, lim_use_2022, liu_identification_2020,liu_multiple_2014, liu_prediction_2020, mao_cross-modality_2021, marti-juan_mc-rvae_2023, sheu_multi-modal_2022, tiwari_multi-modal_2011,vivar_multi-modal_2018, wang_interpretability-based_2022, wang_towards_2023,xin_interpretation_2021, xu_explainable_2021,zhang_tformer_2023}

& not reported \cite{aghili_addressing_2022,li_transfer_2023}; Acc=97\% \cite{dai_-janet_2023}; Acc= 94\% \cite{dong_high-order_2021}; Acc= 88\% \cite{eshaghi_classification_2015};  
Acc= 0.80 \cite{guan_mri-based_2021}; Acc=99.1\% \cite{jabeen_breast_2022}; Acc=99.46\% \cite{khan_stomachnet_2020};  
Acc= 97\% \cite{lei_discriminative_2016}; AUC= 0.86 \cite{liu_identification_2020}; Acc= 91\% \cite{liu_multiple_2014};  
C-index= 0.86–0.94 \newline \cite{liu_prediction_2020}; Acc= 93\% \cite{mao_cross-modality_2021}; Acc= 94\% \cite{marti-juan_mc-rvae_2023};  
Acc= 75\% \cite{sheu_multi-modal_2022}; AUC= 0.90 \cite{tiwari_multi-modal_2011}; AUC= 0.95 \cite{vivar_multi-modal_2018};  
AUC= 0.91 \cite{xu_explainable_2021}; AUC= 0.94 \cite{lim_use_2022}; Acc= 0.91 \cite{wang_interpretability-based_2022};  
AUC= 0.9 \cite{wang_towards_2023}
\\
& Prediction \cite{hatami_cnn-lstm_2022, karaman_machine_2022, kustowski_transfer_2021, li_inferring_2020,shickel_deep_2021, tiulpin_multimodal_2019, wang_knowledge-infused_2022, xu_multi-modal_2022}

& $wR \approx 0.72\text{--}0.80$ \cite{xu_multi-modal_2022}; AUROC= 0.92 \cite{shickel_deep_2021}; \newline Not reported \cite{kustowski_transfer_2021}; AUROC= 0.90 \cite{li_inferring_2020}; \newline AUC= 0.77 \cite{hatami_cnn-lstm_2022}; ROC $\approx$ 0.90 \cite{karaman_machine_2022}; AUC= 0.79 \cite{tiulpin_multimodal_2019}; \newline MAPE= 0.165 \cite{wang_knowledge-infused_2022} \\
& Survival prediction (C-index) \cite{rabinovici-cohen_multimodal_2022,yoo_deeppdt-net_2022} 
& AUC= 0.75 \cite{rabinovici-cohen_multimodal_2022}; Acc= 88\% \cite{yoo_deeppdt-net_2022} \\
& Joint (Class. + Regr.) \cite{thung_identification_2013} 
& Acc= 0.912 \cite{thung_identification_2013} \\
& Generation / Reconstruction \cite{ziegler_multi-modal_2022} 
& Not reported \\
\hline

\rowcolor{groupgray}\multicolumn{3}{|l|}{\textbf{Genomic + Clinical}} \\
& Survival prediction (C-index) \cite{hao_interpretable_2019,hou_hybrid_2023},  
& C-index= 0.63 \cite{hao_interpretable_2019}; C-index= 0.60–0.75 \cite{hou_hybrid_2023}  \\
& Other (Federated learning) \cite{cremonesi_need_2023} 
& Not reported \\
\hline

\rowcolor{groupgray}\multicolumn{3}{|l|}{\textbf{Imaging + Genomic + Clinical}} \\
& Classification (AUC / Acc) \cite{golovanevsky_multimodal_2022,xing_gradient_2023},  
& Acc= 96.9\% \cite{golovanevsky_multimodal_2022}, AUC= 0.92 \cite{xing_gradient_2023}  \\
& Biomarker discovery / diagnostics \cite{boehm_harnessing_2022} 
& Not reported \\
\hline

\rowcolor{groupgray}\multicolumn{3}{|l|}{\textbf{Wearable + Clinical}} \\
& Classification (Acc) \cite{mao_cross-modality_2021} 
& Acc= 93.4\% \cite{mao_cross-modality_2021} \\
& Representation / Feature learning \cite{bahador_deep_2021,xia_graph_2023} 
& Acc= 95\% \cite{bahador_deep_2021}; Acc= 59\% \cite{xia_graph_2023} \\
& Joint (Class. + Regr.) \cite{sukei_automatic_2023} 
& AUPRC= 0.56 \cite{sukei_automatic_2023} \\
& Regression (MAE) \cite{von_luhmann_improved_2020} 
& RMSE reduction=55\% \cite{von_luhmann_improved_2020} \\
\hline
\rowcolor{groupgray}\multicolumn{3}{|l|}{\textbf{Other multimodal (synthetic / simulation / non-medical)}} \\
& Classification (Acc) \cite{wang_multimodal_2020} 
& Acc= 75\%  \\
& Representation learning (Acc) \cite{zhang_feature_2014} 
& Acc= 95\% (ETH-80);  \\
& Prediction (Bias correction) \cite{kustowski_transfer_2021} 
& not reported in standard biomedical metrics \\
& Authentication (Error rates) \cite{tran_raboc_2016}  
& reduced EER by up to 32\% and HTER by 28\% \\
\hline

\rowcolor{groupgray}\multicolumn{3}{|l|}{\textbf{Clinical / EHR}} \\
& Prediction (AUROC) \cite{ahuja_mixehr-guided_2022,an_time-aware_2021,li_inferring_2020}  
& AUROC = 0.84–0.96 \cite{ahuja_mixehr-guided_2022}; 
AUC= 0.94; \newline F1= 0.88 \cite{an_time-aware_2021},  AUROC= 0.85–0.90 \cite{li_inferring_2020}  \\

\hline
\end{tabularx}
\end{table}
We additionally examined reproducibility-related reporting practices across the included studies. A publicly available benchmark dataset (e.g., ADNI, TCGA), as distinct from a private or institutional dataset, was used in 47 of 69 studies (68.1\%), most frequently the Alzheimer's Disease Neuroimaging Initiative (ADNI, 18 studies), The Cancer Genome Atlas (TCGA, 8 studies), and ImageNet (4 studies, primarily for transfer learning rather than direct multimodal evaluation); the remaining studies relied on private or institutional data. Public source code was reported in 23 of 69 studies (33.3\%), most commonly via a public GitHub repository, and only 19 studies (27.5\%) reported both a publicly available benchmark dataset and public source code, while 18 studies (26.1\%) lacked both a publicly available benchmark dataset and public source code (Figure~\ref{fig:fig7}).

Table~\ref{tab:table2} also highlights substantial heterogeneity in the evaluation metrics used across studies. Classification tasks were most commonly assessed using accuracy, AUC, AUROC, sensitivity, specificity, and F1-score, whereas survival prediction studies primarily relied on the concordance index (C-index). Regression and prediction tasks frequently reported RMSE, MAE, MAPE, or task-specific measures. While these metrics are generally appropriate for their respective applications, the diversity of reported outcomes complicates direct comparisons between multimodal approaches. Moreover, differences in datasets, outcome definitions, and validation strategies further limit the ability to determine whether observed performance differences arise from methodological improvements or from variations in evaluation protocols. These observations highlight the need for standardized benchmark datasets and reporting practices to facilitate more meaningful comparisons across multimodal learning studies.
\subsection*{Challenges and solutions}

\subsubsection*{Missing Modalities}
In multimodal data settings, there are typically two types of missing data: the first involves missing values within a modality, where individual modalities contain incomplete information, and the second involves the complete absence of an entire modality for certain samples. This means that for some data points, one type of information might be entirely missing, adding an extra layer of complexity. Addressing this challenge requires specialized imputation approaches distinct from those used for handling missing values within individual modalities. Missing modalities are a frequent issue in medical settings, as various examinations are typically conducted for different patients. Moreover, patients may lack access to certain modalities due to factors such as high costs, safety concerns, sensor malfunction, data corruption, or patient dropout \cite{zhang2022m3care}.

While this systematic review focuses on missing modalities in multimodal data, it's important to acknowledge the broader literature on missing data, which offers foundational approaches for handling such challenges. Traditional methods for handling missing data involve simple imputation methods \cite{little2019statistical}, which replace missing values with mean or median from the known data, and model-based imputation \cite{Ngueilbaye2021} which uses multivariate statistical methods and leverages feature dependencies to impute missing values. Examples of such model-based methods include k-nearest neighbors \cite{liao2014missing}, MissForest (which uses Random Forests) \cite{tang2017random}, and matrix completion methods \cite{fan_polynomial_2020}.
Additionally, multiple imputation methods \cite{rubin_multiple_2018} create several different plausible datasets by filling in missing values multiple times and combining the results to account for the uncertainty of the imputed values.
However, in the context of multimodal datasets, these approaches face limitations. Multimodal datasets contain different types of information from various sources for the same subject, and these approaches do not effectively consider the correlations between these modalities (essential information missed in one modality can be maintained in another modality). For instance, simple imputation methods might oversimplify complex relationships present between modalities, model-based imputation relies on assumptions of linear relationships among modalities, and deep learning methods might not explicitly leverage cross-modality correlations. Since each modality contributes unique insights that need to be considered collectively for accurate imputation, it is necessary to develop techniques that can effectively capture and leverage these interdependencies between different modalities.

Common deep learning-based approaches, such as Variational Autoencoder (VAE)-based and Generative Adversarial Network (GAN)-based methods, provide innovative solutions for missing data imputation. However, they face challenges in capturing multimodal correlations. For instance, the VAE model proposed by \citet{mccoy_variational_2018} can handle missing data but does not account for cross-modality relationships. To address this, \citet{wu2018multimodal} extended the VAE framework to the Multimodal Variational Autoencoder (MVAE), which utilizes a product of expert inference networks to better integrate unique modality contributions \cite{hinton2002training}. However, MVAE struggles to handle randomly missing modalities effectively due to its reliance on fixed input-output structures, limiting its use in clinical applications.

Several studies included in our systematic review highlight the limitations of conventional imputation techniques for multimodal data and propose more advanced approaches for handling missing modalities. 
Missing modalities were addressed in 15 studies, primarily in neurology and oncology applications and most commonly in imaging--clinical and imaging--genomic datasets (Figure~\ref{fig:fig4} and Table~\ref{tab:table2}). The most common solution categories were matrix completion, generative models (VAEs and GANs), graph-based methods, and knowledge distillation (Table~\ref{tab:table1}).

For taking the correlation between modalities into account, \citet{saad_learning-based_2022} proposed using a multitasking learning approach. This approach involves dividing incomplete multimodal datasets into subsets, each containing a different combination of modalities. Individual classifiers are then trained independently on these subsets, and their outputs are combined to generate the final classifier. However, this method assumes linear relationships between modalities, which is often unrealistic.
Another approach, proposed by \citet{vivar_multi-modal_2018}, involves matrix completion techniques by concatenating multimodal features into a single matrix. Yet, this approach fails when entire modalities are missing, as missing entries form blocks rather than random distributions, violating assumptions of matrix completion methods. 
To address these limitations, extensions of VAE and GAN frameworks have been proposed. For example, \citet{xu_explainable_2021} introduced the Dynamic Multimodal Variational Autoencoder (DMVAE), which incorporates modality indicators for missing modalities and leverages joint representations to capture correlations across modalities. This enables DMVAE to impute missing features more effectively than MVAE.  However, DMVAE is not well-suited for image data. In this context, GAN-based approaches provide alternative solutions. \citet{cai_deep_2018} adapted GANs for multimodal settings, using an encoder-decoder architecture to generate missing modalities based on existing ones. To extend this capability beyond imaging, \citet{ziegler_multi-modal_2022} proposed the Multimodal Conditional GAN (MMCGAN). This model combines conditional GANs for tabular data with models for conditional 3D image synthesis, preserving cross-modal correlations and generating realistic synthetic data across diverse modalities. In addition to these generative methods, \citet{hou_hybrid_2023} introduced a Hybrid Graph Convolutional Network that incorporates an online masked autoencoder to leverage cross-modal dependencies via transformers. However \citet{wang_multimodal_2020} highlighted a drawback of imputation-based methods, noting that they can introduce noise, especially when only a limited number of complete modality samples are available, which can negatively affect downstream tasks. To address this, they proposed a novel multimodal learning framework that leverages all available samples, including those with incomplete modalities, without relying on imputation. Their approach is based on knowledge distillation \cite{hinton_distilling_2015}. In this framework, modality-specific models are first trained independently on all available data, even if some modalities are missing. These models act as ``teachers'' to train a multimodal ``student'' model. The student model learns by combining the soft labels provided by the teacher models with the true one-hot labels. Because each teacher model is trained exclusively on a single modality, it benefits from a larger effective sample size, making it a robust expert for its respective modality. The student model integrates these modality-specific insights, enabling it to combine information across all modalities effectively. This approach stands out because it does not discard incomplete modality samples or directly impute missing data. Instead, it ensures that even incomplete samples contribute to the expertise of the teacher models, maximizing data utilization \cite{wang_multimodal_2020}. However, \citet{du_agree_2020} noted that not all knowledge sources are equally reliable, and conflicting information can arise during distillation. Simply weighing all knowledge sources equally can reduce the effectiveness of the student model. To address this limitation, \citet{xing_gradient_2023} proposed the Gradient-guided Knowledge Modulation (GKM) scheme. GKM dynamically adjusts the influence of each knowledge source based on its consistency in the gradient space. By prioritizing reliable information and filtering out misleading knowledge, GKM enhances the distillation process, ensuring the student model absorbs useful insights while minimizing conflicts. This selective adjustment improves the robustness and performance of the student network, offering more dependable guidance from diverse sources.
Overall, the reviewed studies indicate a shift from traditional imputation-based approaches toward representation-learning and knowledge-transfer methods that explicitly exploit cross-modal relationships. Generative models, including VAE- and GAN-based frameworks, together with knowledge-distillation approaches, were among the most frequently adopted solutions. Several studies applying these methods in imaging-based multimodal settings reported strong predictive performance, with classification AUC values of approximately 0.9 or higher in some applications (Table~\ref{tab:table2}). However, differences in datasets, modality combinations, prediction tasks, performance metrics, and evaluation protocols preclude direct comparison across studies, preventing definitive conclusions regarding the superiority of any single approach.

\subsubsection*{Small datasets}

 One of the major challenges in multimodal data fusion is dealing with Small datasets. 
Limited, insufficient, or sparse multimodal data make it difficult to capture the underlying relationships between modalities and to train robust predictive models. Small datasets were addressed in 17 studies, primarily in oncology and neurology applications and most commonly in imaging--clinical and imaging--genomic settings (Figure~\ref{fig:fig4} and Table~\ref{tab:table2}). The most common approaches included data augmentation, transfer learning, knowledge distillation, and multi-task learning (Table~\ref{tab:table1}). 
Researchers have proposed various innovative approaches to address the small data challenge, ranging from data augmentation to transfer learning. Data augmentation techniques \cite{jabeen_breast_2022} stand out, artificially expanding dataset size and diversity, thereby enhancing the training of robust fusion models. Another strategy involves multimodal pre-training, where models learn generic representations from large unimodal datasets, applicable to smaller multimodal datasets. Multi-task learning \cite{saad_dual_2022,liu_prediction_2020} emerges as a powerful technique, enabling the model to jointly train on related tasks, leveraging shared information across modalities to improve performance, even with limited data. Ensemble learning, combining multiple models or techniques, also proves effective in addressing the small data challenge. On the other hand, \citet{cai_multimodal_2022} 
proposed a novel multimodal transformer for image and metadata fusion, incorporating pre-trained Vision Transformer (ViT) models.


In the context of medical imaging, \citet{li_transfer_2023} proposed transfer learning approaches based on fine-tuning convolutional neural networks pre-trained on large natural image datasets to mitigate the scarcity of medical image data. Similarly, \citet{shickel_deep_2021}, \citet{lim_use_2022}, \citet{jabeen_breast_2022}, and \citet{yoo_deeppdt-net_2022} addressed the small sample size challenge using transfer learning. In these approaches, knowledge acquired from models trained on larger datasets is transferred to domain-specific medical tasks through fine-tuning, thereby improving model performance and generalization in settings with limited training data.

However, \citet{lei_discriminative_2016} argued that the ``small-n-large-p'' problem, characterized by a large feature dimension and small sample size, presents persistent challenges in identifying clinical subjects correctly via robust modeling. Both dimension reduction and feature selection offer promising solutions, addressing over-fitting issues and reducing computational time. 
\citet{saad_learning-based_2022},  
taking a different approach to tackle this challenge, they utilize traditional canonical correlation analysis for data fusion highlighting its efficacy in leveraging hidden linear correlation among features that are both small size and heterogeneous. Which can overcome the challenge of small datasets, particularly for 2D images and 1D genomic data. \\
\citet{kustowski_transfer_2021} argued against a purely data-driven approach with limited data and proposed a hybrid methodology combining knowledge-driven and data-driven approaches. The process begins with a simulation-trained neural network surrogate model, which captures crucial correlations among different data modalities and between simulation inputs and outputs. Subsequently, this knowledge is transferred to retrain the model, aligning it with real-world data for improved predictive performance.
In another approach, \citet{guan_mri-based_2021} addressed the challenge of limited data by using knowledge distillation. 
This technique transfers insights from a more complex teacher network to a smaller student network \cite{gou2021knowledge}. In a similar way, the knowledge obtained from multimodal data by a teacher model can be transferred to a student model. Faced with a scarcity of Magnetic Resonance Imaging (MRI) data, \citet{guan_mri-based_2021} propose a multi-modal multi-instance distillation strategy. This approach aims to transfer insights from a multi-modal teacher network to a single-modal student network. The teacher network, trained with both MRI and clinical data, captured comprehensive disease-related patterns. This knowledge was distilled into the student network, which was trained only on MRI data, ensuring that the richer, multi-modal insights were preserved and utilized effectively.
Overall, transfer learning emerged as the most frequently adopted strategy for addressing small-data limitations, particularly in imaging-based multimodal applications where large pre-trained models are available. More broadly, the reviewed studies indicate a shift toward leveraging external sources of information through transfer learning, knowledge distillation, data augmentation, and multi-task learning to improve generalization. Several studies applying these approaches in imaging-based multimodal settings reported strong predictive performance, including AUC values exceeding 0.9 in some classification tasks (Table~\ref{tab:table2}). However, differences in datasets, modality combinations, prediction tasks, and evaluation protocols preclude direct comparison across studies.
\subsubsection*{Interpretability}
Interpretability in multimodal modeling refers to understanding how a model integrates and assigns significance to information from various sources, making its predictions explainable. This is particularly important in medical fields, where decisions made based on model outputs can directly impact patient health. 
Interpretability was addressed in 14 studies across multiple clinical domains, most frequently in oncology and neurology applications (Figure~\ref{fig:fig4} and Table~\ref{tab:table2}). The most common approaches included attention mechanisms, gradient-based attribution methods (e.g., Grad-CAM and integrated gradients), SHAP, feature-importance ranking, and inherently interpretable model architectures (Table~\ref{tab:table1}).
Various techniques have been developed to improve the interpretability of deep learning models in multimodal environments. One such method is the use of attention mechanisms, which allow the model to focus on specific aspects of the data during prediction. For instance, these mechanisms can highlight important regions in a medical image or key elements in genomic sequences, thereby improving transparency in the prediction process.
Another approach involves gradient-based attribution algorithms, which identify and quantify the contribution of each modality to the final prediction. These methods are particularly useful in understanding the biological relevance of the model’s outputs. For instance, to increase the interpretability in a classification task and fusion of metadata and medical imaging data \cite{li_transfer_2023} introduced the ``gradient-weighted class activation mapping'' method, which generates visual explanations by highlighting the regions of the input medical images that are important for the model's prediction.

For the fusion of multi-omic data and clinical data within the context of prognosis prediction in survival analysis, \citet{jiang_autosurv_2024} and \citet{hao_interpretable_2019} addressed the challenge of the black-box nature of deep neural networks in two different ways. 
\citet{jiang_autosurv_2024} 
incorporated Deep SHapley Additive exPlanations (DeepSHAP), a technique that combines deep learning with Shapley values, to explain individual predictions made by the model. The AUTOSurv model, which predicts cancer prognosis, uses DeepSHAP to analyze the SHAP values and quantify how each input feature, such as gene expression or miRNA data, contributes to the final output. This approach enhances the model’s interpretability by linking key pathways with relevant genes, which not only improves the model’s transparency but also highlights potential biomarkers for cancer prognosis.

Similarly, \citet{hao_interpretable_2019} developed CoxPSANet, a deep learning model for cancer survival prediction, which integrates gene expression data at the individual gene level and biological pathways. By establishing sparse connections between the gene and pathway layers, CoxPSANet enhances interpretability by focusing on key biological components and pathways that are relevant to cancer prognosis.
For the fusion of imaging and non-imaging data, \citet{xin_interpretation_2021} proposed the ``Interpretable Deep Multimodal Fusion'' framework, which utilizes Canonical Correlation Analysis to interpret the complex, non-linear associations between modalities. They introduced a cross-modal association (CA) score to assess the impact of each feature, further enhancing the interpretability of the multimodal fusion process.

Overall, the reviewed studies indicate that post-hoc explanation techniques, including SHAP, Grad-CAM, integrated gradients, and feature-importance ranking methods, remain the dominant approaches for improving interpretability in multimodal models. Several studies also explored inherently interpretable architectures that incorporate biological pathways, sparse connections, or explicit cross-modal association measures. Many of these approaches were applied in imaging--clinical and genomic--clinical settings while maintaining strong predictive performance (Table~\ref{tab:table2}). However, differences in datasets, tasks, and evaluation strategies limit direct comparison across studies.

\subsubsection*{Imbalance in dimensionality}
The term ``imbalance in dimensionality'' in the context of multimodal data fusion refers to situations in which the different modalities have significantly different numbers of dimensions or features. An imaging modality, for example, could include millions of pixels, but a genomic modality may have only a few hundred or thousand features.

In data fusion, the dimensional imbalance can be problematic since it might lead to an over-reliance on one modality at the expense of others. In some situations, the high-dimensional modality may dominate the fusion process, resulting in poor low-dimensional modality performance. This can lead to information loss and reduced accuracy in final predictions.
 Several data fusion strategies have been developed to address the challenge of dimensional imbalance. One approach is to use intermediate fusion strategies, in which modality-specific representations are first learned separately before being fused. This allows the representations to be projected into comparable latent spaces, reducing the influence of large differences in feature dimensionality across modalities. Another option is to lower the dimensionality of the high-dimensional modality using feature selection or dimensionality reduction algorithms. This can help to prevent the high-dimensional modality from dominating the fusion process and increasing the low-dimensional modality's performance.
\\
Seven studies addressed imbalance in dimensionality, primarily in neurology, oncology, and physiology applications. This challenge was most commonly investigated in imaging--clinical and imaging--genomic settings (Figure~\ref{fig:fig4} and Table~\ref{tab:table2}). The most common approaches included feature selection, dimensionality reduction, weighted and focal loss functions, and architectures designed to balance information across modalities (Table~\ref{tab:table1}).
For example, \citet{karaman_machine_2022} and \citet{partin_data_2023} employed weighted loss functions to assign greater importance to underrepresented modalities or classes, counteracting the overemphasis on high-dimensional data. Similarly, \citet{mao_cross-modality_2021} and \citet{li_transfer_2023} utilized focal loss functions to concentrate learning on harder-to-classify samples, which helped reduce bias toward dominant modalities and enhanced model robustness.
Other approaches included non-linear dimensionality reduction, as demonstrated by \citet{madabhushi_computer-aided_2011}, which aimed to preserve the intrinsic structure of the data while minimizing dimensionality-related imbalances. In a different vein, \citet{xu_multi-modal_2022} leveraged recurrent neural networks to model temporal dependencies, helping to better align and integrate multimodal inputs over time.

Although imbalance in dimensionality focuses on feature-level disparities, related issues such as class imbalance, particularly common in high-dimensional spaces, can further complicate multimodal classification tasks. \citet{tran_raboc_2016} addressed this by introducing the RABOC algorithm, a hybrid method combining one-class classification and Real AdaBoost. One-class classification is inherently suited to extreme class imbalance, as it defines decision boundaries using only the target class, while Real AdaBoost boosts performance by focusing on misclassified samples through weighted re-training. Together, these methods effectively handle severe class imbalance in high-dimensional multimodal biometric authentication tasks without overfitting.
Similarly, \citet{li_transfer_2023} tackled class imbalance in multimodal classification through the Deep Multimodal Generative Adversarial Network (DMGAN). This approach generates synthetic features for minority classes via adversarial training, thereby rebalancing the dataset. DMGAN outperforms traditional oversampling techniques like SMOTE (Synthetic Minority Over-sampling Technique) \cite{Alex2023}, XGBoost (eXtreme Gradient Boosting)  \cite{chen_scalable_2016}, Unimodal GAN frameworks \cite{Sauber-Cole2022}, and Random Forest \cite{More2017}, demonstrating the effectiveness of generative approaches in dealing with imbalanced multimodal data.
Overall, the reviewed studies indicate that feature-selection and dimensionality-reduction approaches remain the most direct strategies for mitigating disparities in feature dimensionality across modalities. Weighted- and focal-loss frameworks further help prevent multimodal models from overemphasizing dominant modalities or classes. Several of these approaches achieved strong classification and prediction performance in imaging-based multimodal settings (Table~\ref{tab:table2}). However, differences in datasets, modality combinations, and evaluation protocols limit direct comparison across studies.
\subsubsection*{Optimal fusion strategy}
To make the most out of multimodal data, an optimal fusion strategy is required.
The optimal fusion strategy in modeling multimodal data depends on various factors such as the specific task requirements, the characteristics of the modalities involved (e.g., the number of modalities, their relationships, dimensionality, and sampling rates), the available computational resources, trade-offs between model complexity, interpretability, performance, and the ease of implementation \cite{ramachandram_deep_2017}. Additionally, the selection of an optimal fusion strategy should also consider the quality and reliability of each modality's data, as well as the level of correlation between the modalities \cite{wang_towards_2023}. By considering all these factors, the most effective fusion strategy can be determined for specific multimodal data modeling tasks. Overall, an optimal fusion strategy in modeling multimodal data involves integrating and combining multiple modalities in a way that maximizes the benefits of each modality while minimizing the biases and limitations inherent in individual modalities \cite{pawlowski_effective_2023}.

In addressing the challenge of finding the optimal fusion strategy, the problem in the context of deep multimodal learning can be reduced to two central architectural design choices. The first choice relates to the level at which different modalities should be combined, which can be guided by considering the challenges and advantages of early, late, and intermediate fusion approaches. The second architectural design choice concerns which modalities to combine, as including all available modalities does not always enhance the performance of a machine learning algorithm \cite{ramachandram_deep_2017}. 
Finding an optimal fusion strategy was identified as a challenge in 13 studies, primarily in oncology and neurology applications. This challenge was most commonly investigated in imaging--clinical and imaging--genomic settings (Figure~\ref{fig:fig4} and Table~\ref{tab:table2}). The most common approaches included attention mechanisms, transformer-based architectures, adaptive weighting, hierarchical fusion, latent representation learning, and correlation-based fusion methods (Table~\ref{tab:table1}).
For example, \citet{xu_mufasa_2021} tackled this challenge by extending the neural architecture search \cite{yao1999evolving,real2019regularized}, which was originally designed for unimodal data, to incorporate the joint optimization of fusion strategies and modality-specific architectures for multimodal data. This extension allows for searching for unique architectures for distinct modalities and identifies the best strategies to fuse these architectures at the right representation level, and then they utilize state-of-the-art neural architecture search methods to explore and optimize fusion strategies for multimodal data. By jointly optimizing fusion strategies and modality-specific architectures, their approach aims to enhance prediction performance on EHR tasks while controlling for computation costs.

In the fusion of imaging and clinical data, \citet{wang_towards_2023} pointed out a common heuristic approach to data fusion in many studies, where data from multiple sources are simply concatenated without proper preprocessing or assessment of fusion quality. This failure to interpret the fused feature space can result in suboptimal model performance, particularly in medical imaging problems where the features of different data sources can vary significantly. Therefore, they proposed modeling the fully connected layer of a deep neural network as a potential function following the classical Gibbs measure. By considering the features of the fully connected layer as random variables governed by state functions representing different data sources, the method quantifies source bias and aims to optimize data fusion quality. The use of a vector-growing encoding scheme, such as positional encoding, helps the conversion of low-dimensional clinical features into a rich feature space to complement high-dimensional imaging features, ultimately enhancing the fusion process in multi-source deep learning problems. \\
Several studies have highlighted that different modalities often have unequal importance and contributions, necessitating innovative methods to adaptively weight modalities during the fusion process. For instance, \citet{dai_-janet_2023} employs dual encoders and joint attention mechanisms to assign greater importance to features that are most relevant for Alzheimer's disease classification. Similarly, \citet{hatami_cnn-lstm_2022} introduces a weighted ensemble method for predicting stroke outcomes, where weights are assigned to imaging module predictions based on their alignment with clinical metadata, thereby improving prediction accuracy. In another approach, \citet{rahaman_deep_2023} uses attention mechanisms to prioritize the most informative modalities, mitigating performance degradation that arises from forcing equal contributions. Lastly, \citet{xia_graph_2023} incorporates an adaptive fusion layer to dynamically assign modality-specific weights, optimizing their contributions in clustering tasks. Collectively, these studies underscore the critical role of adaptive weighting techniques in enhancing the effectiveness of multimodal fusion by accounting for the varying importance of different modalities.
In the classification task, there are various approaches to determining the optimal fusion strategy. For example, \citet{pawlowski_effective_2023} emphasized the significance of selecting the most appropriate fusion technique. Specifically in classification tasks, they outlined three criteria for making design decisions when selecting the most suitable data fusion technique. These criteria involve evaluating the influence of each modality on the machine learning problem, considering the task type, and assessing memory usage during both training and prediction phases. \\
There are also two similar works by \citet{khan_stomachnet_2020} and \citet{jabeen_breast_2022} which determine the optimal fusion strategy in two steps. Firstly, they select the best features using the differential evolution selected features.
\citet{khan_stomachnet_2020}  introduced a probability-based fusion method that involves computing probabilities for selected features and then choosing one feature based on the highest probability value. This selected feature is used as a reference for comparison, and features from different vectors are merged into a single matrix. This comparison step helps address the issue of redundant features in the vectors. 
On the other hand, \citet{jabeen_breast_2022} proposed the maximum correlation-based fusion approach, aiming to combine optimal feature vectors by selecting features with strong correlations, discarding those with weak correlations, and merging them into a single vector. By prioritizing features with high correlation coefficients, this fusion method enhances the representational power of the fused vector. This approach ensures that the combined features effectively capture complementary information, leading to improved classification accuracy.
Overall, the reviewed studies suggest a shift from static fusion schemes toward adaptive and attention-based strategies that dynamically adjust the contribution of individual modalities. Transformer-based architectures, attention mechanisms, and adaptive weighting were among the most frequently proposed solutions. Several studies applying these approaches in imaging--clinical and imaging--genomic settings reported strong predictive performance (Table~\ref{tab:table2}). However, differences in modality combinations, datasets, prediction tasks, and evaluation metrics preclude direct comparison across studies.
\subsubsection*{Other challenges}
Beyond the main challenges detailed in this manuscript, there is a range of unique additional issues that do not align with predefined categories yet are still critical for understanding and improving multimodal data integration. In Supplementary Table S3, the specific challenges for ``Other Challenges'' are explicitly stated alongside their corresponding solutions, as they vary across different studies. One major challenge is extracting high-order interactions within and among data modalities \cite{dong_high-order_2021,zhang_feature_2014}, which is crucial for identifying complex, non-linear relationships that may exist between features. These interactions are particularly relevant in clinical contexts where modalities such as genomics and imaging provide complementary insights. Similarly, capturing the intricate relationships and temporal dynamics across modalities \cite{an_time-aware_2021,brand_joint_2020}, while accounting for irregularities in patient visit timings, is a significant hurdle. Patient data often exhibits temporal patterns, with varying time gaps between visits, making it essential to develop models capable of handling such inconsistencies.
High dimensionality adds another layer of complexity \cite{bahador_deep_2021,liu_identification_2020} as many modalities, such as imaging and genomics, have a vast number of features. This can lead to intensive computational requirements and the risk of overfitting, necessitating robust dimensionality reduction and feature selection techniques. Capturing intermodality interactions \cite{dai_-janet_2023,krix_multigml_2023,golovanevsky_multimodal_2022,an_time-aware_2021,mao_cross-modality_2021,alam_kernel_2018,zille_enforcing_2017} further complicates modeling, as it requires mechanisms that can integrate and prioritize information from heterogeneous sources without losing critical cross-modal correlations. Lastly, heterogeneity in data \cite{rabinovici-cohen_multimodal_2022,saad_learning-based_2022,abdelaziz_alzheimers_2021,dong_high-order_2021,li_inferring_2020,eshaghi_classification_2015,madabhushi_computer-aided_2011}, whether in terms of formats, scales, or quality, presents a persistent challenge, particularly in ensuring that diverse modalities are integrated cohesively. These challenges highlight the need for innovative approaches, such as advanced attention mechanisms, hybrid fusion strategies, and efficient computational frameworks, to enable robust and interpretable multimodal data integration in clinical applications.

\section{Discussion} \label{sec:discussion}
In this systematic review, we identified several key technical challenges that hinder the modeling of multimodal medical data and assessed how recent studies have attempted to overcome them. The most commonly addressed obstacles were missing modalities and small sample sizes, which were tackled in 15 (22\%) and 17 (25\%) of the 69 studies, respectively.

A range of innovative solutions have been proposed to address these challenges. For missing modalities, researchers have used multitask learning \cite{saad_learning-based_2022}, matrix completion \cite{thung_identification_2013,fan_polynomial_2020}, variational autoencoders \cite{marti-juan_mc-rvae_2023,wu2018multimodal}, generative adversarial networks \cite{ziegler_multi-modal_2022,cai_deep_2018}, hybrid graph convolutional networks \cite{hou_hybrid_2023}, and knowledge distillation \cite{wang_multi-modal_2023}, all aiming to impute or compensate for absent modalities. For limited sample sizes, strategies such as data augmentation \cite{partin_data_2023,jabeen_breast_2022}, transfer learning from large pretrained models \cite{sukei_automatic_2023,li_transfer_2023,jabeen_breast_2022,lim_use_2022,shickel_deep_2021,yoo_deeppdt-net_2022}, and knowledge distillation \cite{guan_mri-based_2021} have been used to maximize the value of scarce data. However, many of these techniques rely on external pretrained models or large datasets that are not always available, and may fail to capture complex cross-modal patterns when data are sparse or heterogeneous. For interpretability, studies have applied attention mechanisms, gradient-based attribution tools \cite{krix_multigml_2023,li_transfer_2023}, and explanation techniques such as DeepSHAP \cite{rabinovici-cohen_multimodal_2022,sheu_multi-modal_2022,xu_explainable_2021,wang_interpretability-based_2022}. To address imbalance in dimensionality across modalities, methods such as feature selection, dimensionality reduction \cite{madabhushi_computer-aided_2011}, and weighted loss functions \cite{karaman_machine_2022,partin_data_2023,mao_cross-modality_2021,li_transfer_2023} have been used. Finally, selecting an optimal fusion strategy remains an open challenge: early, intermediate, late, and hybrid fusion each have distinct strengths and weaknesses, and advances such as neural architecture search and attention-based fusion \cite{dai_-janet_2023,rahaman_deep_2023,vanguri_multimodal_2022} show promise for tailoring integration to specific tasks.
Each of these approaches has inherent limitations. Methods for handling missing modalities often rely on assumptions about inter-modality relationships and may be sensitive to noisy or incomplete data; transfer learning and augmentation depend on suitable source data; and advanced fusion architectures increase computational complexity. No single methodological solution is universally applicable, the best strategy depends on the data and the clinical task.

These approaches also differ substantially in computational requirements. While transfer learning can reduce training costs by leveraging pretrained models, methods based on GANs \cite{ziegler_multi-modal_2022,cai_deep_2018}, variational autoencoders \cite{marti-juan_mc-rvae_2023}, graph-based architectures \cite{hou_hybrid_2023,krix_multigml_2023}, and attention-based fusion \cite{dai_-janet_2023,zhang_tformer_2023} may require greater resources and longer training times, a consideration particularly relevant when integrating high-dimensional modalities such as imaging, genomics, and longitudinal records, and one that must be balanced against predictive performance in real-world deployment.

Notably, these challenges are often interconnected rather than independent. Missing modalities are particularly problematic in small datasets, where limited samples reduce a model's ability to learn robust cross-modal relationships and increase overfitting risk \cite{wang_multimodal_2020,guan_mri-based_2021}. Imbalance in dimensionality can similarly be exacerbated in data-scarce settings, as high-dimensional modalities dominate learning while lower-dimensional ones contribute little, a challenge related to the small-\emph{n}, large-\emph{p} problem \cite{lei_discriminative_2016}. These challenges can also compound one another: methods that reconstruct missing modalities often need additional parameters and training data, potentially worsening small-data limitations \cite{hou_hybrid_2023}. Conversely, some methods address multiple challenges at once, knowledge distillation can leverage incomplete samples while improving efficiency in limited-data settings \cite{wang_multimodal_2020,guan_mri-based_2021}, and intermediate/attention-based fusion can reduce imbalance in dimensionality while remaining more robust to missing modalities than early fusion. Future methodological work should therefore increasingly target integrated solutions addressing multiple challenges simultaneously.

While this review does not introduce new methods, it systematically synthesizes existing approaches by the specific modeling challenge they address, rather than by data modality or algorithmic technique as in prior reviews. To our knowledge, this is the first review to categorize multimodal modeling methods by the challenges they address across diverse medical applications, offering practical guidance for method selection and highlighting areas where development remains limited.
Notably, only 14 of 69 included studies explicitly addressed interpretability. Given the growing complexity of multimodal deep learning, future work should prioritize multimodal-specific explainability methods, better understanding of cross-modal interactions, and standardized frameworks for evaluating the clinical usefulness of explanations.

The small-data challenge, though frequently mentioned, remains difficult to overcome; even the most common approaches, augmentation and transfer learning, have inherent limitations when data are extremely scarce or unrepresentative. Several included studies reported improved performance via transfer learning from larger datasets \cite{li_transfer_2023,lim_use_2022,shickel_deep_2021,yoo_deeppdt-net_2022}, particularly in imaging applications with readily available pretrained models, while others used data augmentation to increase training diversity and reduce overfitting \cite{partin_data_2023,jabeen_breast_2022}. Effectiveness, however, varied with source-data availability, source–target domain similarity, and modality characteristics. 
Multicenter knowledge-transfer strategies that leverage information across clinical sites offer another promising direction for addressing data scarcity \cite{Wang2024KnowledgeTransfer}. Few-shot learning, which enables adaptation from very few labeled examples, was rarely represented in the included studies but may also be valuable where large annotated datasets are hard to obtain.

No single fusion strategy emerged as universally preferred, consistent with observations by \citet{snoek_early_2005}, who found that the relative performance of early versus late fusion depends on the task \cite{zhang_feature_2014}. Studies on optimal fusion strategy were most common in oncology and neurology, combining imaging, genomic, and clinical data. Intermediate and hybrid fusion were often favored for enabling modality-specific feature extraction while preserving cross-modal interactions; late fusion was more common with heterogeneous or partially missing modalities; and early fusion, while simpler, is harder to apply when modalities differ substantially in dimensionality or structure. Several imaging–clinical and imaging–genomic studies used attention mechanisms, adaptive weighting, and transformer-based architectures to dynamically weight modality contributions \cite{dai_-janet_2023,rahaman_deep_2023,xia_graph_2023}. Differences in datasets, modality combinations, tasks, and metrics make direct comparisons difficult, and no strategy consistently outperformed others.

Developments outside the clinical domain reinforce the relevance of these challenges. In emotion recognition, sentiment analysis, and intelligent transportation, generative modality-dropout frameworks have improved robustness to partially missing modalities \cite{zhang2025generative}; tensor-decomposition-based fusion has reduced parameter complexity and overfitting in high-dimensional representations \cite{wang2024multimodal}; hierarchical and cross-attention-based fusion has improved modeling of heterogeneous modality interactions \cite{zhu2024emotion}; and robust adversarial fusion frameworks have improved resilience to noisy or incomplete modalities \cite{wang2025raft}. These developments suggest that missing modalities, imbalance in dimensionality, and fusion optimization are broader methodological issues, not unique to healthcare, and that other domains may inspire future clinical methodological advances.

Beyond the deep learning approaches represented in the included studies, the broader literature has explored statistical and geometric frameworks for multimodal integration, including multimodal subspace learning and support-vector-data-description approaches such as MSSVDD \cite{sohrab2021multimodal}, manifold-alignment techniques \cite{nguyen2022deep}, modality-specific representation learning and regularization \cite{yu2021learning}, and low-rank representation methods. 
Recent entropy-regularized multimodal subspace learning frameworks further improve robustness to dimensional imbalance and small sample sizes through adaptive weighting and shared subspace representations \cite{Wang2025EntropySVDD}. 
These methods aim to learn shared cross-modal representations while preserving complementary modality-specific information and reducing redundancy; although underrepresented in our included studies, they offer complementary directions for future clinical applications, particularly with limited samples, heterogeneous sources, or high-dimensional features.

Emerging paradigms such as prompt-based learning, contrastive learning, and knowledge-transfer approaches have shown promise in applications like depression recognition and sentiment analysis \cite{liu2025prompt,zhang2024contrastive,wang2024crossmodal,li2024depression}. Prompt-based learning guides multimodal feature extraction and fusion via task-specific prompts and domain knowledge \cite{liu2025prompt}; contrastive learning improves cross-modal representation and alignment by encouraging consistency across heterogeneous sources while preserving modality-specific information \cite{zhang2024contrastive,wang2024crossmodal}; and knowledge-transfer/representation-learning frameworks improve modeling when labeled data are limited \cite{li2024depression}. Though largely explored outside the clinical domain so far, these approaches are promising for addressing modality alignment, fusion, and data scarcity in future clinical applications.

This study has several limitations. First, our review may not capture all existing methodologies in this fast-evolving field. 
The search was conducted using PubMed and Google Scholar, which offer broad biomedical and interdisciplinary coverage but may have missed studies indexed exclusively in computer-science-oriented databases such as IEEE Xplore, ACM Digital Library, Scopus, or arXiv; future reviews may benefit from broader database coverage. Second, the effectiveness of the discussed solutions is context-dependent; what works for one dataset or problem may not generalize to others. Third, despite our efforts at comprehensive coverage, some studies or viewpoints may have been missed.
Fourth, our search ended in October 2023, so newer developments are not captured. Recent work has moved toward generalizable, foundation-model approaches integrating diverse data types, clinical records, imaging, pathology, and molecular data \cite{Soenksen2022HAIM,Tripathi2025HONeYBEE}. HAIM combined tabular, time-series, text, and imaging data in a unified, modular pipeline with Shapley-value analysis of modality contributions \cite{Soenksen2022HAIM}, and HONeYBEE generated unified patient-level embeddings while handling missing modalities and flexible fusion \cite{Tripathi2025HONeYBEE}. Yet these studies confirm many of our review's challenges persist: HONeYBEE found clinical embeddings often outperformed multimodal ones, with fusion helping only for specific tasks and cancer types, underscoring that multimodal integration's value is context-dependent \cite{Tripathi2025HONeYBEE}. Newer multimodal large language models and multi-agent systems such as BrainGPT and MAM extend this toward clinical reasoning and report generation but still face evaluation, generalizability, and implementation challenges \cite{Li2025BrainGPT,Zhou2025MAM}. In short, missing modalities, interpretability, robustness, and clinical deployment remain open problems as the field evolves.
Fifth, although we initially highlighted chronic diseases such as COPD and asthma as motivating examples given their multimodal nature, our search yielded very few studies explicitly focused on these conditions. This may reflect true underrepresentation of such diseases in current multimodal modeling research rather than a search-strategy failure, since our queries were designed to capture a broad range of clinical domains; we highlight this as an important area for future research. Finally, although our search was not limited to deep learning-based approaches, most included studies used deep learning, so the review's focus on deep learning is a result of the evidence base rather than by design.


Several methodological limitations of the review process should also be acknowledged. This review was not prospectively registered (e.g., in PROSPERO), which may reduce transparency despite following predefined eligibility criteria, a predefined search strategy, and PRISMA guidelines. Inter-rater agreement statistics (e.g., Cohen's kappa) were not calculated, although each full-text study was independently reviewed by two reviewers with disagreements resolved by a third. Finally, a formal quality or risk-of-bias appraisal was not performed, as the substantial methodological heterogeneity of included studies precluded use of a single established tool; findings should therefore be interpreted as a qualitative synthesis of methodological approaches rather than a comparative quality assessment.

Beyond these methodological issues, broader concerns of fairness, ethics, and deployment are increasingly recognized as critical in multimodal AI for medicine. Several challenges discussed in this review may also affect fairness and equity: small datasets can limit representation of diverse patient populations, and missing modalities may disproportionately affect groups for whom certain data sources are less consistently available. Fairness research remains scarce in many clinical domains, with sensitive attributes such as race, sex, and socio-economic status often underrepresented in available datasets, limiting equity in model performance \cite{liu2023translational,norori2023fairness}. In medical imaging, algorithms can exploit demographic shortcuts that perform well on benchmarks but fail to generalize to new populations \cite{yang2024limits}, and the multimodal setting adds complexity as different modalities may carry distinct, potentially reinforcing, sources of bias \cite{hickmon2024multimodal}. Beyond fairness, clinical deployment raises concerns including data privacy, informed consent for integrating heterogeneous sources, interpretability for clinicians, and alignment with regulatory frameworks.

Relatedly, the data landscape underlying the included studies was dominated by a small number of large, well-established public datasets. Across the 69 included studies, the Alzheimer's Disease Neuroimaging Initiative (ADNI) was the most frequently used public benchmark (18 studies), combining imaging, clinical, and cognitive data for Alzheimer's disease and mild cognitive impairment research. The Cancer Genome Atlas (TCGA) was the second most common (8 studies), spanning multiple cancer subtypes for integrating genomic, pathological, and clinical data, while ImageNet was used in 4 studies, primarily for transfer learning rather than direct multimodal evaluation. Other domain-specific benchmarks, including MIMIC-III and Derm7pt, appeared less frequently. Overall, 47 of 69 studies (68.1\%) used at least one publicly available benchmark dataset, while the remainder relied on private or institutional data. This concentration around a few large public datasets reflects their comprehensiveness and central role in enabling reproducible research, but also highlights an important opportunity for the field to develop and adopt additional standardized, publicly available multimodal benchmarks across a broader range of clinical domains, which would in turn support the kind of rigorous external validation needed for clinical translation.
Software and code availability were similarly limited across the included studies. Only 23 of the 69 included studies (33.3\%) reported publicly available source code, most commonly via a public GitHub repository, while the remaining 46 studies (66.7\%) did not provide any publicly accessible code. Only 19 studies (27.5\%) reported both public source code and use of a publicly available benchmark dataset, meaning the large majority of included studies could not be readily reproduced or extended by other researchers using the information provided in the publication alone. Greater adoption of open-source practices, alongside the standardized benchmarks discussed above, would substantially improve the reproducibility, transparency, and downstream adoption of multimodal modeling approaches in clinical research.

Successful clinical translation requires rigorous external validation across diverse populations and settings, as well as prospective evaluation and real-world implementation studies to assess clinical utility, workflow integration, and robustness under routine conditions. Regulatory approval processes increasingly require evidence of safety, reliability, transparency, and ongoing performance monitoring, particularly for adaptive AI systems that evolve over time; while beyond the scope of this review, these represent important directions for ensuring that technical advances translate into safe, equitable, and clinically useful applications.
Future research should prioritize integrated methods that address multiple interconnected challenges simultaneously. Small datasets frequently compound other challenges, exacerbating missing-modality effects and dimensionality imbalance, so progress on small-data-specific techniques (e.g., incorporating domain-specific knowledge, improving data simulation, or developing hybrid models that combine data-driven approaches with clinical insight) is likely to yield benefits across several challenges at once. Adaptive fusion strategies robust to missing modalities and multimodal-specific interpretability methods represent complementary priorities for advancing the field.

\section{Conclusion} \label{sec:conclusion}

In this systematic review, we synthesized evidence from 69 studies and identified five recurring methodological challenges in multimodal medical data modeling: missing modalities, small datasets, imbalance in dimensionality, interpretability, and the selection of appropriate fusion strategies. Although numerous approaches have been proposed to address these challenges, no single methodological approach consistently performs best across all clinical applications. Instead, the effectiveness of existing solutions depends on the characteristics of the available data, modality combinations, prediction task, and clinical context.

Our findings further suggest that these challenges should not be viewed in isolation. Missing modalities, limited sample sizes, imbalance in dimensionality, interpretability, and fusion strategy selection frequently interact, indicating that future methodological developments should increasingly focus on integrated approaches capable of addressing multiple challenges simultaneously. While considerable progress has been made in areas such as handling missing modalities and developing advanced fusion techniques, important methodological gaps remain. These include improving multimodal interpretability, developing robust learning methods for data-scarce clinical settings, designing adaptive fusion strategies, and ensuring reliable generalization across diverse healthcare settings.

Future research should therefore prioritize multimodal methods that are not only accurate but also robust, interpretable, computationally efficient, and clinically transferable. This includes the development of multimodal-specific explainability techniques, adaptive fusion strategies, learning frameworks for limited-data settings, and rigorous external validation across diverse patient populations. Addressing these priorities will be essential for translating advances in multimodal learning into safe, reliable, and clinically useful healthcare applications.

Overall, this review provides a challenge-oriented synthesis of the current literature that complements existing modality- and application-focused reviews by systematically mapping recurring methodological challenges to the solutions proposed to address them while identifying common methodological patterns, remaining limitations, and priorities for future methodological development. We hope that this perspective will help researchers identify appropriate strategies for specific multimodal modeling problems while guiding future methodological advances in multimodal medical data modeling.

\section*{Abbreviations}
\small
\begin{tabular}{@{}ll@{}}
AI & Artificial Intelligence \\
AUC & Area Under the Curve \\
AUROC & Area Under the Receiver Operating Characteristic Curve \\
AUPRC & Area Under the Precision--Recall Curve \\
EHR & Electronic Health Record \\
GAN & Generative Adversarial Network \\
MAE & Mean Absolute Error \\
MAPE & Mean Absolute Percentage Error \\
MRI & Magnetic Resonance Imaging \\
MVAE & Multimodal Variational Autoencoder \\
RMSE & Root Mean Square Error \\
ROC & Receiver Operating Characteristic \\
VAE & Variational Autoencoder \\
CT & Computed Tomography \\
COPD & Chronic Obstructive Pulmonary Disease \\
PRISMA & Preferred Reporting Items for Systematic Reviews and Meta-Analyses \\
DMVAE & Dynamic Multimodal Variational Autoencoder\\
DMGAN & Deep Multimodal Generative Adversarial Network\\
SHAP  & SHapley Additive exPlanations \\
GKM & Gradient-guided Knowledge Modulation \\
RABOC    &  Real AdaBoost and One-Class classification (hybrid algorithm) \\

\end{tabular}
\normalsize

\section*{Supplementary Information}
The Supplementary Material for this article includes:
Supplementary Table S1. Extracted data from the 69 included studies, provided as a separate Excel file. The table contains information on clinical domains, fusion strategies, data modalities, challenges and corresponding solutions, task types, and performance metrics.
Supplementary Table S2. PubMed search query used to identify relevant articles (search date: 2023/10/09).
Supplementary Table S3. Summary of challenges and solutions identified across the 69 included studies, organized by study publication.
The Supplementary Material can be found online.

\section*{Acknowledgements}
We thank Daniela Karakuleva for her assistance with the extraction of data modalities and performance metrics.
\section*{Author contributions} 
MF, MW, HBi, and NB contributed to the study's conception and design. MF, MW, MB, JK, and ChG were involved in the systematic search process. NH extracted the medical domain and disease type of the articles. MF extracted the data, while MW reviewed the extraction. SW contributed expertise on the state of the art in multimodal modeling research in computer science. MF wrote the initial draft of the manuscript, with NB providing supervision throughout the process. HBa contributed to the interpretation of the results and the revision of the manuscript. All authors provided critical feedback and approved the final version.
\section*{Funding} The work of MF, SW, HBa, HBi, and NB has been funded in part by the Deutsche Forschungsgemeinschaft (DFG, German Research Foundation) – Project-ID 499552394 – SFB 1597. The work of MW, ChG, MB, JK, and HBi was partially funded by the Federal Ministry of Education and Research (BMBF) – CALM-QE project, under grant numbers 01ZZ2318B, 01ZZ2318L, and 01ZZ2318I.

\section*{Data Availability}
All data supporting the findings of this study are available within the paper and its Supplementary Material.

\section*{Declarations}
\subsection*{Ethics approval and consent to participate} Not applicable
\subsection*{Consent for publication} Not applicable
\subsection*{Competing interests}
The authors declare no competing interests

\bibliography{References}

\end{document}